\documentclass[review,3p]{elsarticle}

\usepackage{amssymb}
\usepackage{amsmath}
\usepackage{booktabs}
\usepackage{multirow}
\usepackage{algorithm}
\usepackage{algorithmic}
\usepackage[hypertexnames=false]{hyperref}
\hypersetup{pdfproducer={},pdfcreator={}}
\usepackage{tabularx}
\usepackage{placeins}
\usepackage{float}

\journal{Transportation Research Part C: Emerging Technologies}

\begin{document}

\begin{frontmatter}

\title{Observation-Aligned Two-Stage Domain Decomposition for Physics-Informed Traffic State Estimation with Sparse Fixed Sensors}

\author[purdue]{Eunhan Ka}
\author[eiffel]{Ludovic Leclercq}
\author[purdue]{Satish V. Ukkusuri\corref{cor1}}
\cortext[cor1]{Corresponding author}
\ead{sukkusur@purdue.edu}

\affiliation[purdue]{organization={Lyles School of Civil and Construction Engineering, Purdue University},
            addressline={610 Purdue Mall},
            city={West Lafayette},
            postcode={47907},
            state={IN},
            country={United States}}

\affiliation[eiffel]{organization={Univ Gustave Eiffel, ENTPE, LICIT-ECO7},                        city={Lyon},
            postcode={F-69675},
                        country={France}}

\begin{abstract}
Traffic state estimation from sparse fixed sensors is challenging because physics-informed neural networks (PINNs) tend to over-smooth sharp transitions admitted by the Lighthill--Whitham--Richards (LWR) model. This study proposes Two-Stage Domain Decomposition Physics-Informed Neural Networks (TSDD-PINN), an observation-aligned framework for LWR-based offline speed-field reconstruction. The framework supports spatial, temporal, and space--time refinement. Matched direction analysis shows that spatial refinement has the lowest mean error and less than half the training time of space--time refinement in the tested setting, while temporal refinement is faster. A global parent PINN is first trained. In the controlled spatial implementation, its residual profile guides a deterministic partition for warm-started child networks. An optional operational safeguard retains Stage~1 when the prespecified screen does not activate. The primary I-24 MOTION evaluation spans five days, five sensor configurations, and ten seeds per configuration, yielding 1{,}500 runs. Controlled TSDD-PINN attains the lowest relative $L_2$ error in 18 of 25 configurations and 14 of 15 sparse-sensing cases, while training 2.4 times faster than the extended PINN (XPINN) baseline under the evaluated implementations and training budgets. Non-neural comparisons show that the advantage over interpolation is concentrated under sparse sensing, whereas dense sensing often favors interpolation. A separate 250-run operational evaluation finds infrequent activation and motivates the Stage-1-preserving safeguard. The residual is treated as an indicator of model difficulty rather than a validated shock detector. The evidence supports a sensing-density-dependent operating range rather than uniform improvement.
\end{abstract}

\begin{highlights}
\item Observation-aligned domain decomposition targets fixed-sensor under-observation.
\item Spatial refinement has lowest mean error and under half the space--time training time.
\item TSDD-PINN has lower error than XPINN in 18/25 cases and is 2.4 times faster.
\item Benefits over linear interpolation concentrate under sparse sensing.
\item A 250-run evaluation motivates a Stage-1-preserving no-trigger safeguard.
\end{highlights}


\end{frontmatter}

\section{Introduction}\label{sec:introduction}

Traffic state estimation (TSE) is the process of inferring complete spatiotemporal fields of traffic variables (density, flow, and speed) from partial observations collected by fixed sensors such as loop detectors and cameras \citep{seo2015probe, seo2017traffic}. Accurate TSE underpins real-time traffic management, traveler information systems, and transportation infrastructure planning. In practice, sensor deployment is constrained by cost and maintenance: a typical freeway corridor may be instrumented with only a small number of detector stations spaced several miles apart, yielding sparse spatial coverage of the roadway. Reconstructing the full traffic state from such limited observations remains a fundamental challenge, particularly during incidents and rapidly evolving congestion.

Physics-based models provide a principled foundation for TSE. The Lighthill-Whitham-Richards (LWR) model \citep{lighthill1955kinematic, richards1956shock}, a first-order hyperbolic conservation law, describes the spatiotemporal evolution of traffic density through mass conservation and a flow-density fundamental relation. Once closed by a fundamental relation, the LWR model predicts kinematic wave propagation, including the formation and dissipation of shockwaves at congestion boundaries. Model-based data assimilation methods, such as extended Kalman filtering \citep{wang2005real, wang2008real}, combine these macroscopic traffic models with real-time sensor data to reconstruct traffic states. Variational and Hamilton-Jacobi formulations provide a complementary line of exact or optimization-based estimation for heterogeneous sensing settings \citep{daganzo2006variational, canepa2017networked}. However, LWR and its higher-order extensions such as the Aw-Rascle-Zhang (ARZ) model \citep{aw2000resurrection, zhang2002} rely on idealized assumptions that limit fidelity under real-world conditions. Data-driven approaches, including deep learning models for traffic speed prediction and imputation \citep{ma2015long, wu2018}, can learn complex spatiotemporal patterns but require large volumes of high-quality training data and provide no guarantee of physical consistency.

\begin{figure}[!htbp]
    \centering
    \includegraphics[width=\textwidth]{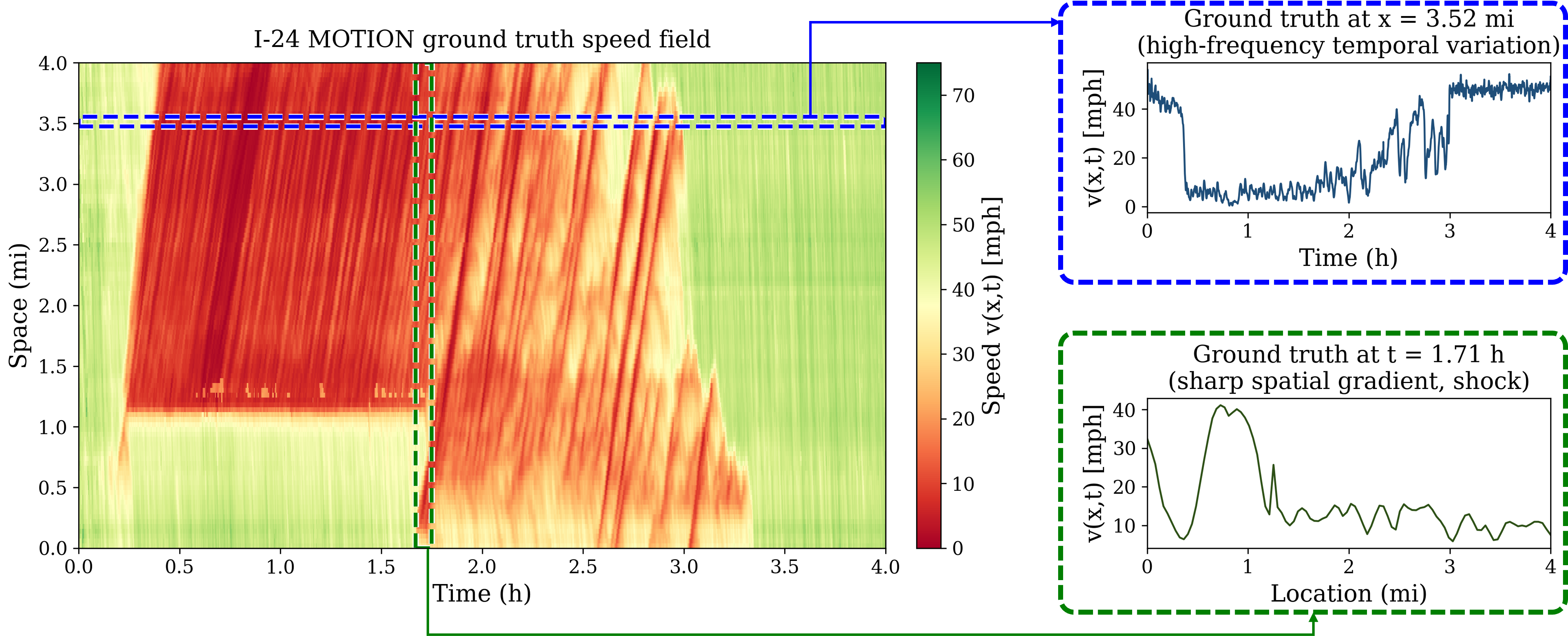}
    \caption{Ground-truth speed field $u(x,t)$ from I-24 MOTION on 2022-11-21 (left), with a temporal trace at $x = 3.52$ mi (top right) and a spatial profile at $t = 1.71$ h (bottom right). Congested regions appear in red and free-flow regions in green.}
    \label{fig:motivation}
\end{figure}

Physics-informed neural networks (PINNs) address this trade-off by embedding the governing partial differential equations (PDEs) directly into the neural network's training objective \citep{raissi2019physics}. The resulting model simultaneously fits available sensor data and satisfies the traffic flow PDE at collocation points throughout the domain, enabling physically consistent estimation even in regions with no direct observations. Several studies have demonstrated the promise of PINNs and related physics-informed learning methods for freeway TSE with the LWR and ARZ models \citep{shi2021a, shi2021b, yuan2021traffic, huang2022physics}. Despite these advances, standard PINNs face several challenges when applied to traffic flow. First, neural networks exhibit \emph{spectral bias}, a tendency to learn smooth, low-frequency functions, limiting their ability to represent the sharp discontinuities characteristic of LWR solutions \citep{tancik2020fourier}. Second, balancing the data loss and PDE loss during training is difficult; inappropriate weighting biases the model toward either underfitting the physics or underfitting the data \citep{wang2021understanding}. Beyond these optimization challenges, the strong-form PDE residual is ill-defined at discontinuities. Pointwise residual minimization is structurally incompatible with the shock solutions admitted by hyperbolic conservation laws such as LWR. These structural difficulties are visible in real freeway data. Figure~\ref{fig:motivation} shows a representative ground-truth speed field from the I-24 MOTION dataset, with sharp spatial gradients across shock fronts and high-frequency temporal variation throughout the congested window. This localized multi-scale structure can be difficult for a single global PINN to resolve under sparse fixed-sensor supervision. \citet{lei2025potential} recently formalized these failure modes, showing that PINNs can perform worse than both their data-driven and physics-based counterparts when spatiotemporal resolution is insufficient or gradient imbalance is severe.

Domain decomposition can reduce local approximation complexity and may mitigate spectral bias: partitioning the spatiotemporal domain into smaller subdomains, each handled by a dedicated sub-network, reduces the local function complexity that any single network must represent. The extended PINNs (XPINNs) framework \citep{jagtap2020extended} generalizes this idea to arbitrary PDEs by decomposing the domain in both space and time and enforcing residual-based continuity at subdomain interfaces. We adopt XPINN as our DD-PINN baseline because it supports spatial, temporal, and space-time decomposition within a unified framework, enabling systematic comparison of decomposition directions. Residual-based adaptive refinement (RAR) complements domain decomposition by concentrating collocation points in high-error regions \citep{wu2023comprehensive}. However, canonical DD-PINN frameworks and many subsequent variants still depend on user-designed or initialization-dependent decompositions \citep{jagtap2020extended, moseley2023finite, hu2023augmented}. More recent adaptive approaches have begun to relax this assumption \citep{peng2025aspinns, si2025idpinn, luo2025pdd, botvinick2025abpinns}; however, these methods target generic PDE benchmarks rather than sparse fixed-sensor traffic state estimation, and they do not study the decomposition direction under fixed-sensor observation geometry. The choice of decomposition direction (spatial, temporal, or space-time) has received limited attention for fixed-sensor traffic state estimation, even though the answer depends strongly on the observation structure and the underlying PDE.

\citet{huang2023limitations} demonstrated the shockwave challenge explicitly for traffic, showing that physics-informed learning on the hyperbolic LWR equation produces errors substantially larger than on a diffusion-augmented counterpart when solutions contain shockwaves. This result motivates methods that resolve discontinuities without sacrificing the hyperbolic character of the conservation law. It also motivates approaches that localize shock-prone transition regions rather than relying on a single global network, artificial smoothing, or direct tracking of a moving shock manifold.

Motivated by these limitations, we develop Two-Stage Domain Decomposition Physics-Informed Neural Networks (TSDD-PINN), an observation-aligned two-stage domain-decomposition framework that first trains a global parent PINN and then transfers its representation to direction-specific child networks when refinement is enabled. The framework is instantiated and evaluated with spatial, temporal, and space--time decompositions. Among the evaluated directions, spatial refinement attains the lowest mean error and requires less than half the training time of space--time refinement, while temporal refinement is faster. In the controlled spatial implementation, the residual profile indicates where the Stage~1 model has difficulty representing the traffic state and guides a deterministic spatial partition. It is not interpreted as a validated shock detector. When the prespecified screen does not activate, an optional operational safeguard retains the Stage~1 predictor. We make four contributions:

\begin{enumerate}
\item \textbf{Observation-aligned two-stage domain decomposition.} {TSDD-PINN} first trains a corridor-wide parent and then, when refinement is enabled, transfers its representation to direction-specific child networks. In the controlled spatial implementation, the residual profile supplies a deterministic partition heuristic for the warm-started child networks. Parent-to-child warm start and fixed-interface regularization are integrated into the same traffic-specific workflow.

\item \textbf{Decomposition-direction analysis under fixed-sensor observation.} A controlled comparison spanning 180 evaluated direction-specific results, comprising 120 newly trained temporal and space--time runs and 60 spatial results reused from the controlled benchmark, evaluates spatial, temporal, and space--time refinement under matched conditions. Among the evaluated directions, spatial decomposition attains the lowest mean error and requires less than half the training time of space--time decomposition, while temporal decomposition is faster. This result does not constitute a general rule for selecting the decomposition direction.

\item \textbf{Controlled accuracy--efficiency evidence.} Under the evaluated implementations, subdomain structures, and training budgets, the combined TSDD-PINN design yields lower mean error and trains 2.4 times faster than the evaluated XPINN baseline; component controls do not attribute this method-level result to Stage~1, initialization, or any single mechanism.

\item \textbf{Multi-day validation and a sensing-density-dependent operating scope.} A 1{,}500-run I-24 evaluation, comparisons with canonical time-slice spatial linear interpolation and the adaptive smoothing method (ASM), both using the same selected virtual-sensor locations, a separate default-screen evaluation, and an NGSIM negative control show that the supported advantage is concentrated under sparse sensing and does not extend uniformly to dense sensing or the NGSIM negative control.
\end{enumerate}

The benchmark includes canonical time-slice spatial linear interpolation and ASM. Linear interpolation is matched to both the observation and evaluation grids, whereas ASM is matched to the selected sensor locations and evaluated on its native time grid. Under the present equally spaced interior-speed virtual-sensor protocol, a matched implementation of extended Kalman filtering or variational Hamilton--\allowbreak Jacobi reconstruction would require additional observations or model specifications, including measured boundary or ramp flows when available, cumulative-count constraints, and calibrated state-transition and measurement models. Both approaches have established applications under observation settings that include such information, and no superiority claim is made over them.

The remainder of this paper is organized as follows. Section~\ref{sec:litreview} reviews relevant literature on TSE methods, PINNs, domain-decomposition PINNs, and PINNs for conservation laws. Section~\ref{sec:methodology} presents the TSDD-PINN methodology and analytical motivation, including the two-stage training framework, operational screening and residual-based partitioning, interface conditions, and decomposition-direction analysis. Section~\ref{sec:experiments} describes the datasets, baseline methods, and evaluation process. Section~\ref{sec:results} reports the quantitative results. Section~\ref{sec:discussion} discusses the findings, transportation implications, and limitations. Section~\ref{sec:conclusion} concludes the paper.

\section{Literature Review}\label{sec:litreview}

\subsection{Traffic State Estimation Methods}\label{sec:lit-tse}

Traffic state estimation methods are commonly categorized as model-based, data-driven, physics-informed, or hybrid.

\paragraph{Model-based approaches} Classical model-based TSE relies on macroscopic traffic flow models combined with filtering, smoothing, or variational inference. The LWR conservation law \citep{lighthill1955kinematic, richards1956shock}, closed by an empirical fundamental relation, remains a common backbone of physics at the corridor scale. The Cell Transmission Model \citep{daganzo1994cell} provided a computationally practical, discrete analog of the LWR that reproduces queues and shockwave propagation, and became a core tool for freeway TSE, filtering, and control. Real-time estimation has relied heavily on extended Kalman filtering and related assimilation methods that combine macroscopic process models with sensor observations \citep{wang2005real, wang2008real}. \citet{treiber2002reconstructing} introduced an adaptive smoothing method for detector-based spatiotemporal traffic reconstruction, which has become a benchmark in the field. Variational and Hamilton-Jacobi formulations provide a complementary line of work by exploiting cumulative-count representations and semi-analytic solution structures to enable exact or optimization-based data assimilation with heterogeneous observations \citep{daganzo2006variational, canepa2017networked}. \citet{seo2017traffic} provide a comprehensive survey classifying TSE approaches by data source, state variable, and estimation paradigm. While model-based methods are physically grounded, they depend on calibration quality and typically assume a fixed or low-dimensional traffic law that may be insufficient to capture the complexity of real traffic dynamics.

\paragraph{Data-driven approaches} Much of the traffic machine learning literature has focused on prediction or imputation rather than direct reconstruction of complete traffic fields. Recurrent neural networks, such as long short-term memory models and hybrid deep architectures, can learn spatiotemporal dependencies from historical sensor data \citep{ma2015long, wu2018}. More directly related to TSE, \citet{rempe2022estimation} used deep convolutional neural networks to estimate continuous speed fields from sparse freeway data. These methods can capture nonlinear traffic patterns directly from observations, but they require representative training data and provide no guarantee of conservation, physically admissible wave propagation, or robustness in unobserved regions.

\paragraph{Physics-informed and hybrid approaches} Physics-informed neural networks (PINNs) and related physics-informed deep learning (PIDL) methods aim to combine the data efficiency of physical models with the representational power of neural networks. \citet{huang2022physics} demonstrated that embedding the LWR PDE into the training objective significantly improves estimation accuracy under sparse sensing. \citet{shi2021a} and \citet{shi2021b} applied physics-informed learning to both first-order and second-order traffic models, showing that shockwave formation and dissipation patterns can be reconstructed from limited data. \citet{barreau2021physics} used physics-informed learning with pretraining to reconstruct traffic density from low-penetration probe vehicles. Beyond standard PINNs, \citet{lu2023physics} and \citet{zhang2024physics} coupled traffic-flow models with differentiable computational graphs to jointly estimate corridor states, queue profiles, and fundamental-diagram parameters, while \citet{zhao2024observer} combined a PDE observer with deep learning for boundary-sensing TSE. \citet{ka2024physics} further extended physics-informed estimation to network-scale settings.

Within the physics-informed traffic literature, domain decomposition has received limited attention. \citet{usama2022physics} extended PINNs to traffic networks by decomposing a network into individual links and enforcing flow conservation constraints at junctions. While this is the closest prior work to decomposition in traffic PINNs, the decomposition is topology-driven, one sub-network per road link, and it does not evaluate spatial, temporal, and space--time decomposition directions within a single freeway corridor under fixed-sensor observation geometry. \citet{huang2023limitations} showed that physics-informed learning for the hyperbolic LWR equation yields substantially larger errors than for a diffusion-augmented counterpart when solutions contain shockwaves, establishing that the shockwave problem has measurable consequences for traffic estimation accuracy and motivating shock-oriented improvements beyond standard single-domain formulations.

\subsection{Physics-Informed Neural Networks: Training Challenges}\label{sec:lit-pinn}

Physics-informed machine learning has emerged as a paradigm for integrating governing equations with data-driven models \citep{karniadakis2021physics}. The PINN framework, formalized by \citet{raissi2019physics}, showed that governing PDEs can serve as a strong inductive bias in small-data regimes by penalizing PDE residuals at collocation points alongside data-fitting objectives. Since then, the primary research challenge has shifted from formulation to trainability.

\emph{Spectral bias} is one such difficulty: neural networks trained with gradient descent learn low-frequency components of the solution faster than high-frequency ones, leading PINNs to over-smooth sharp gradients and discontinuities \citep{tancik2020fourier}. Fourier feature embeddings, which project the low-dimensional input into a higher-dimensional sinusoidal space, have been shown to mitigate this bias by enabling the network to represent high-frequency content from early in training.

A second challenge is \emph{gradient pathology} in the composite PINN loss. \citet{wang2021understanding} showed that imbalance among data, boundary, and PDE loss gradients can substantially slow or destabilize convergence. \citet{krishnapriyan2021characterizing} demonstrated systematic PINN failures on convection-dominated and reaction-dominated PDEs, arguing that soft PDE regularization can become ineffective on more challenging problems. \citet{lei2025potential} recently formalized these failure modes specifically for traffic flow, showing that PINNs can perform worse than both their data-driven and physics-based counterparts when spatiotemporal resolution is insufficient or when gradient imbalance is severe.

Several remedies have been proposed. \citet{wang2024respecting} introduced causal training, which respects the temporal structure of time-dependent PDEs by weighting PDE residuals according to their position in the causal sequence. \citet{wu2023comprehensive} developed residual-based adaptive refinement (RAR), which dynamically adds collocation points in high-error regions. Self-adaptive loss weighting \citep{mcclenny2023self} treats loss weights as learnable parameters, though the effectiveness of such schemes varies across problem settings \citep{lei2025potential}. These remedies improve optimization within a single global network but do not resolve the structural incompatibility between pointwise residual minimization and discontinuous solutions of hyperbolic conservation laws. This limitation motivates domain decomposition approaches that reduce the complexity of local functions.

\subsection{Domain Decomposition PINNs}\label{sec:lit-ddpinn}

Domain decomposition (DD) reduces the complexity that any single network must represent. By partitioning the spatiotemporal domain into smaller subdomains, each handled by a dedicated sub-network, DD-PINNs represent local features more accurately and enable parallel training.

\citet{jagtap2020conservative} introduced conservative PINNs (cPINNs), which decompose the domain spatially and enforce strong-form flux continuity across interfaces, making the method particularly well-suited to conservation laws. \citet{jagtap2020extended} generalized this idea into the extended PINN (XPINN) framework, which supports arbitrary space-time decompositions with residual-based continuity at subdomain interfaces. We adopt XPINN as our DD-PINN baseline because it supports spatial, temporal, and space-time decomposition within a unified framework, enabling systematic comparison of decomposition directions. \citet{shukla2021parallel} developed a parallel implementation of cPINNs and XPINNs, clarifying the computational trade-offs: cPINNs are communication-efficient for spatial decomposition, while XPINNs are more flexible because they also support temporal splitting. \citet{hu2021extended} analyzed when XPINNs improve generalization, showing that decomposition creates a tradeoff between simpler local functions and fewer samples per subdomain. \citet{moseley2023finite} proposed finite basis PINNs (FBPINNs), which use overlapping subdomains with compact-support basis functions to improve scalability and reduce spectral bias. \citet{hu2023augmented} introduced augmented PINNs (APINNs), which employ a gating network to realize soft, trainable domain decomposition with flexible parameter sharing.

Prior DD-PINN frameworks and many subsequent variants still depend on user-designed or initialization-dependent decompositions. More recent adaptive approaches have begun to relax this assumption. \citet{peng2025aspinns} proposed AS-PINNs, which adapt interface positions through adversarial interaction among subnetworks for discontinuous high-order equations. \citet{si2025idpinn} proposed IDPINN, which uses a coarse small-data PINN to initialize subdomain networks and augments interface smoothness losses. Unlike IDPINN, which focuses on initialization-enhanced decomposition for generic PDE benchmarks, the present study evaluates decomposition direction, fixed-sensor traffic-state reconstruction, and sensing-density-dependent operating scope. \citet{luo2025pdd} introduced progressive domain decomposition (PDD), which segments the domain according to residual-loss dynamics and preserves successful local models. \citet{botvinick2025abpinns} developed AB-PINNs, which insert new basis subdomains in regions of high residual loss during training. However, these methods have been developed for generic PDE benchmarks rather than for sparse fixed-sensor traffic state estimation, and they do not examine the direction of decomposition under fixed-sensor observation geometry. Their interface designs are also not tailored to LWR-based freeway estimation, where conservation and shock-prone transitions are central.

Thus, the remaining gap is an observation-aligned two-stage domain-decomposition framework that relates decomposition direction to fixed-sensor observation geometry and evaluates spatial, temporal, and space--time alternatives under matched traffic-reconstruction conditions.

\subsection{PINNs for Conservation Laws with Shockwaves}\label{sec:lit-shocks}

Hyperbolic conservation laws, such as the LWR, admit weak solutions with discontinuities (shockwaves), posing a fundamental challenge for PINNs that minimize a pointwise strong-form PDE residual. At a shock, the classical PDE residual is undefined because the solution gradient diverges, and neural networks, as continuous function approximators, cannot accurately represent the discontinuous solution.

\citet{huang2023limitations} demonstrated this difficulty explicitly for traffic, showing that physics-informed learning on the first-order hyperbolic LWR equation produces substantially larger errors than on a diffusion-augmented counterpart when solutions are non-smooth. This finding provides the most direct traffic-specific motivation for the present work. Several lines of research have sought to address the shockwave challenge. \citet{coutinho2023physics} added adaptive localized artificial viscosity to stabilize PINNs near shocks, trading physical fidelity for numerical tractability. \citet{deryck2024wpinns} proposed weak PINNs (wPINNs), which replace the strong-form residual with a weak formulation suited to entropy solutions of scalar conservation laws. \citet{neelan2025physics} studied the gap between conservative and non-conservative PDE formulations in PINNs, clarifying when Rankine-Hugoniot conditions must be explicitly enforced. \citet{lorin2024nondiffusive} proposed a non-diffusive neural network method that constructs smooth local solutions in subdomains separated by discontinuity lines defined from Rankine-Hugoniot jump conditions, thereby learning the discontinuity geometry itself. For adaptive sampling, \citet{lu2021deepxde} introduced RAR in the DeepXDE framework, and \citet{wu2023comprehensive} showed that residual-based adaptive sampling significantly improves accuracy near shocks compared to static collocation.

Most shock-oriented PINN models remain single-formulation methods based on viscosity, weak losses, or adaptive sampling, whereas explicit discontinuity-tracking methods such as non-diffusive neural networks (NDNNs) learn the discontinuity geometry itself. TSDD-PINN does not attempt to track the moving shock manifold directly; instead, it uses the coarse residual as an indicator of model difficulty and as a deterministic heuristic for spatial partitioning, then applies regularization at fixed interfaces. This positioning differs from weak-form and discontinuity-line methods and does not treat residual peaks or valleys as validated shock locations. An observation-aligned coarse-to-fine domain decomposition workflow, combined with matched evaluation across spatial, temporal, and space-time directions, has received limited attention in sparse fixed-sensor TSE.

In summary, the reviewed literature identifies four challenges for sparse, fixed-sensor freeway TSE. Alternative decomposition schemes require transparent indicators derived from the traffic estimation problem itself; the decomposition direction should be matched to fixed-sensor observation geometry, and coarse-to-fine warm starts remain underdeveloped for traffic DD-PINNs. Multi-day real-world validation with statistical testing is also needed to distinguish systematic gains from dataset-specific or seed-specific effects.

\section{Methodology}\label{sec:methodology}

\begin{figure}[!htbp]
    \centering
    \includegraphics[width=\linewidth]{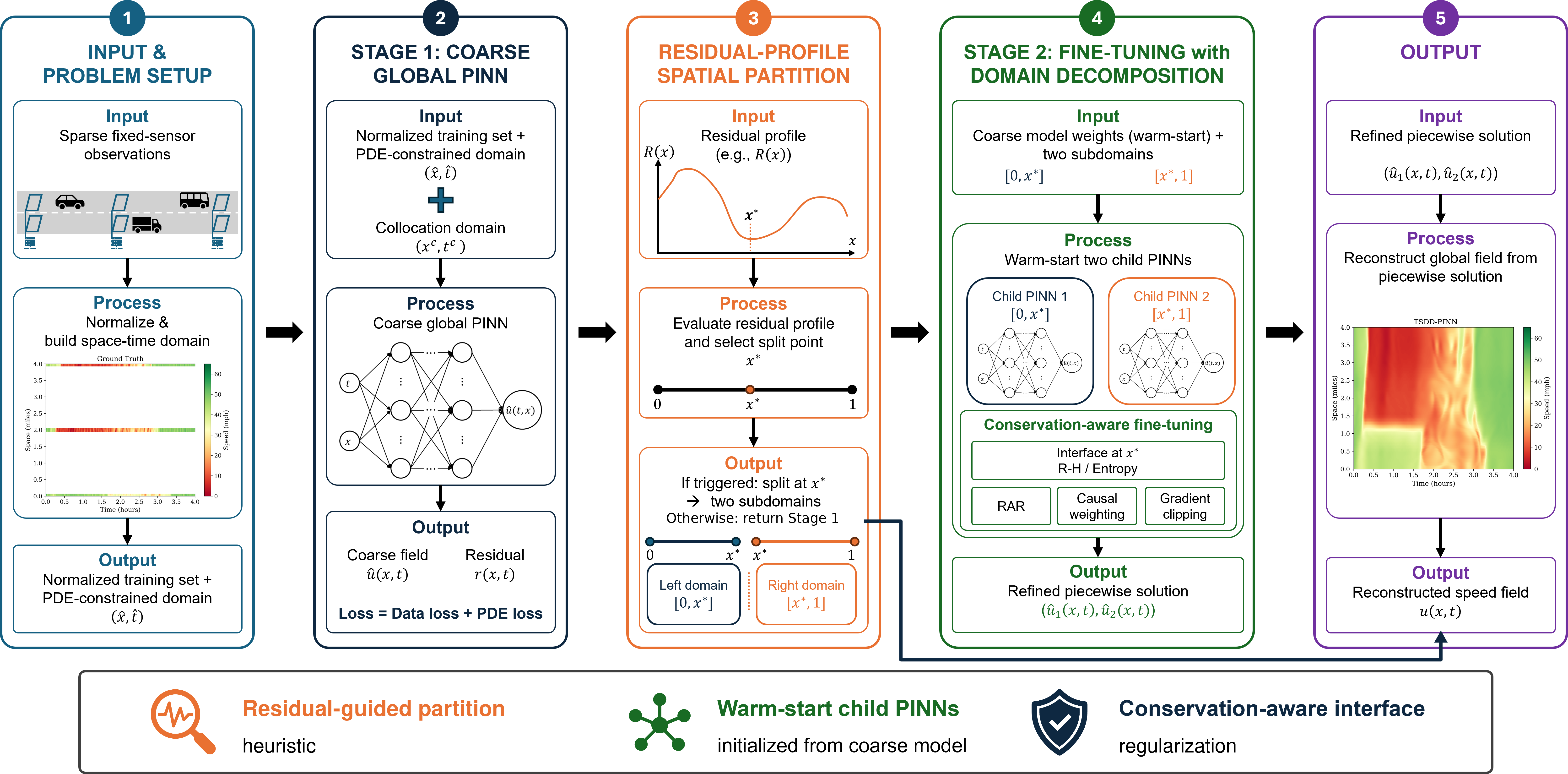}
    \caption{Framework overview of {TSDD-PINN}. Stage~1 is common; the controlled path refines, while the operational path applies the prespecified screen and the optional no-trigger safeguard returns captured Stage~1 when the screen does not activate. This safeguard does not validate the screen. Figure~2 depicts the spatial implementation used in the controlled benchmark; matched temporal and space--time variants are defined in Section~\ref{sec:direction} and evaluated in Section~\ref{sec:results-ablation}.}
    \label{fig:framework}
\end{figure}

This section presents the {TSDD-PINN} framework for traffic state estimation, illustrated in Figure~\ref{fig:framework}. We first formulate the problem (Section~\ref{sec:problem-formulation}), then describe the network architecture (Section~\ref{sec:network-arch}), loss function (Section~\ref{sec:loss}), two-stage training framework (Section~\ref{sec:two-stage}), operational screening and residual-based partitioning (Section~\ref{sec:adaptive-detection}), interface conditions (Section~\ref{sec:interface}), additional training components (Section~\ref{sec:additional-components}), and decomposition direction analysis (Section~\ref{sec:direction}).

\subsection{Problem Formulation}\label{sec:problem-formulation}

We consider the problem of reconstructing the complete speed field $u(x,t)$ across a freeway segment from sparse fixed-sensor observations. This is a PDE-constrained state reconstruction (data assimilation) problem: unlike classical forward problems, no initial or boundary conditions are specified; unlike classical inverse problems, no PDE parameters are inferred. In the primary offline benchmark, $v_f$ is the 95th percentile of the complete trajectory-derived aggregated speed field and is held fixed throughout training.

Let $\Omega = [x_{\min}, x_{\max}] \times [0, T]$ denote the spatiotemporal domain. Traffic dynamics are governed by the Lighthill-Whitham-Richards (LWR) conservation law \citep{lighthill1955kinematic, richards1956shock}:
\begin{equation}\label{eq:lwr-density}
\frac{\partial \rho}{\partial t} + \frac{\partial q(\rho)}{\partial x} = 0,
\end{equation}
where $\rho(x,t)$ is the traffic density and $q(\rho)$ is the flow. In this study, we close Equation~\ref{eq:lwr-density} with the Greenshields fundamental diagram \citep{greenshields1935study}:
\begin{equation}\label{eq:greenshields}
q(\rho) = v_f \, \rho \left(1 - \frac{\rho}{\rho_{\mathrm{jam}}}\right),
\end{equation}
with $v_f$ the free-flow speed and $\rho_{\mathrm{jam}}$ the jam density.
This choice should be interpreted as one admissible first-order closure rather than a defining feature of {TSDD-PINN}. Triangular diagrams are a standard alternative in freeway kinematic-wave modeling \citep{newell1993queueing}. We use Greenshields here because its smooth, single-valued, differentiable form yields the speed-form residual in Equation~\ref{eq:lwr-speed} without piecewise branches or a min operator. A triangular or calibrated closure would require replacing the smooth residual and the interface flux or characteristic terms with the corresponding piecewise or min-operator expressions. The residual-guided split detection and parent-to-child warm start do not depend on the parabolic form of Equation~\ref{eq:greenshields}, because they operate on residual evaluations and network weights. Using the relation $u = v_f(1 - \rho/\rho_{\mathrm{jam}})$, the conservation law can be rewritten in speed form:
\begin{equation}\label{eq:lwr-speed}
\frac{\partial u}{\partial t} + (2u - v_f)\frac{\partial u}{\partial x} = 0.
\end{equation}

Both speed and spatial/temporal coordinates are normalized to $[0,1]$ via min-max scaling before being input to the neural network:
\begin{equation}\label{eq:normalization}
\hat{x} = \frac{x - x_{\min}}{x_{\max} - x_{\min}}, \qquad \hat{t} = \frac{t}{T}, \qquad \hat{u} = \frac{u - u_{\min}}{u_{\max} - u_{\min}},
\end{equation}
where $u_{\min}$ and $u_{\max}$ are the physical speed bounds obtained from the complete aggregated speed field derived from trajectories, not from the selected sparse virtual sensor traces.
All subsequent equations in Sections~\ref{sec:network-arch} to \ref{sec:direction} operate in normalized coordinates unless otherwise noted. In this system, $\hat{\rho}_{\mathrm{jam}} = \hat{u}_{\max} = 1$. We work consistently in these normalized coordinates throughout training, so the normalized density $\hat{\rho} = 1 - \hat{u}$ used in Section~\ref{sec:interface} is a normalized analog of the physical density rather than the physical density; the Rankine-Hugoniot and entropy losses are optional conditional interface regularizers in normalized coordinates rather than exact mass-conservation statements in physical units. Conversion to physical units uses $u = u_{\min} + \hat{u}(u_{\max} - u_{\min})$ and the corresponding flux scaling. After normalization, the PDE residual takes the form
\begin{equation}\label{eq:residual}
r(x,t) = \frac{A \dfrac{\partial \hat{u}}{\partial x} \;-\; B\,\hat{u}\,\dfrac{\partial \hat{u}}{\partial x} \;-\; \dfrac{\partial \hat{u}}{\partial t}}{\sqrt{A^2 + B^2 + 1}},
\end{equation}
where $\hat{u}(x,t;\theta)$ is the network prediction in normalized coordinates and the coefficients are derived from the normalization:
\begin{equation}\label{eq:nondim-coeffs}
C = c_{\mathrm{fps}} \cdot \frac{T_{\mathrm{range}}}{X_{\mathrm{range}}}, \qquad
A = (v_f - 2\,u_{\min}) \cdot C, \qquad
B = 2\,(u_{\max} - u_{\min}) \cdot C.
\end{equation}
Here, $c_{\mathrm{fps}} = 5280/3600$ converts miles per hour to feet per second, $T_{\mathrm{range}}$ and $X_{\mathrm{range}}$ are the temporal and spatial extents of the domain in seconds and feet respectively, and $u_{\min}$, $u_{\max}$ follow the complete-field definition above. The denominator $\sqrt{A^2 + B^2 + 1}$ rescales the coefficient vector $(A,-B,-1)$ to unit Euclidean norm; it does not guarantee an order-one residual magnitude.

A set of $N_{\mathrm{data}}$ sensor observations $\mathcal{O} = \{(x_i^d, t_i^d, u_i^{\mathrm{obs}})\}_{i=1}^{N_{\mathrm{data}}}$ is available from fixed-location detectors at interior spatial positions. The TSE problem is formulated as the following optimization:
\begin{equation}\label{eq:optimization}
\theta^* = \arg\min_\theta \mathcal{L}_{\mathrm{total}}(\theta),
\end{equation}
where the total loss $\mathcal{L}_{\mathrm{total}}$ is defined in Section~\ref{sec:loss}.
The formulation assumes a single connected freeway corridor without on-ramp or off-ramp source terms and uses lane-aggregated traffic states. These assumptions define the scope of the present evaluation and keep the comparison focused on domain decomposition rather than on traffic-model complexity. This formulation differs from classical forward PDE problems, which seek to propagate a solution from prescribed initial and boundary conditions, and from inverse problems, which seek to infer unknown PDE parameters from observations. Here, the network is an empirically evaluated regularized estimator of the interior field from sparse traces and the prescribed LWR residual with fixed $v_f$. Since the benchmark also uses complete-field normalization statistics and virtual sensors, it represents offline reconstruction rather than a deployment-ready inverse problem.

Let
\(\Omega=[x_{\min},x_{\max}]\times[0,T]\), with
\(x_{\min}<x_{\max}\) and \(T>0\), and let
\(S=\{x_1,\ldots,x_{n_s}\}\), indexed so that
\(x_1<\cdots<x_{n_s}\), denote the \(n_s\) fixed-sensor locations.
Let \(C^1(\Omega)\) denote the space of continuously differentiable
speed fields on \(\Omega\). Let \(\mathcal{H}_S\) denote the
fixed-sensor temporal-trace operator. For a speed field
\(u\in C^1(\Omega)\), its output is
\begin{equation}\label{eq:obs-operator}
\mathcal{H}_S u
=
\bigl(u(x_1,\cdot),\ldots,u(x_{n_s},\cdot)\bigr),
\end{equation}
where the dot \(\cdot\) denotes the time variable and
\(u(x_i,\cdot)\) is the complete temporal trace at sensor \(x_i\).
\ref{app:theory-proofs} constructs infinitely many linearly
independent field variations that produce zero traces at every sensor
for all times.

For two adjacent sensors at \(x_i\) and \(x_{i+1}\), let
\(h=x_{i+1}-x_i>0\) denote the sensor gap.
At a fixed time, the speed profile on this
interval is noiseless and constant at \(u_L\) up to a single
discontinuity, beyond which it is constant at \(u_R\). The signed
change in speed across the discontinuity is \(\Delta u = u_R - u_L\),
and the discontinuity lies at \(x_i + \alpha h\), where
\(\alpha \in (0,1)\) gives its relative position within the interval.
The two sensors observe \(u_L\) and \(u_R\), and \(I_h u\)
denotes the linear function joining these values.
Then
\begin{equation}\label{eq:interp-bound}
\left\|u-I_hu\right\|_{L^2(x_i,x_{i+1})}^2
=
\frac{(\Delta u)^2h}{3}
\left[\alpha^3+(1-\alpha)^3\right]
\geq
\frac{(\Delta u)^2h}{12}.
\end{equation}
The norm is evaluated on the interval between adjacent sensors. Its dependence on \(h\) indicates greater interpolation error over wider unobserved intervals and motivates the empirical comparison of refinement directions. This is neither an error bound for TSDD-PINN nor a theorem on direction selection.
\ref{app:theory-proofs} provides the assumptions, proof, and
fixed-split representation identity.

\subsection{Network Architecture}\label{sec:network-arch}

Each subdomain $\Omega_s$ is assigned a dedicated feedforward neural network $\mathrm{NN}_s(x,t;\theta_s)$. The normalized input coordinates $(x,t)$ first pass through a Fourier feature mapping \citep{tancik2020fourier}:
\begin{equation}\label{eq:fourier}
\gamma(x,t) = \bigl[\sin(\mathbf{W}[x,t]^\top),\; \cos(\mathbf{W}[x,t]^\top)\bigr] \in \mathbb{R}^{2d_e},
\end{equation}
where $\mathbf{W} \in \mathbb{R}^{d_e \times 2}$ is a random matrix drawn from $\mathcal{N}(0, \sigma^2)$ with $\sigma = 10$ at initialization and held fixed thereafter. The embedding dimension $d_e$ is set to half the first hidden layer width: for a first hidden layer of 256 neurons, $d_e = 128$, producing a 256-dimensional Fourier embedding. This mapping projects the two-dimensional input into a higher-dimensional space of sinusoidal features, enabling the network to learn high-frequency components that would otherwise be suppressed by spectral bias.

The Fourier features feed into a sequence of fully connected hidden layers with $\tanh$ activations, followed by a linear output layer that produces a single scalar: the predicted speed. The $\tanh$ activation is used because its smoothness keeps the derivatives required by the PDE residual continuous. Prior to domain decomposition, the parent network uses the architecture $[2, 256, 128, 128, 128, 1]$, where the first dimension denotes the raw input size before Fourier embedding and the subsequent dimensions are the widths of the hidden and output layers. After decomposition, each child network uses $[2, 256, 128, 128, 1]$, reducing the depth by one hidden layer to balance expressiveness with computational cost while retaining the same Fourier embedding dimension.

The layer counts should be interpreted as fixed implementation hyperparameters rather than calibrated optima. Single-domain methods use the parent architecture, while decomposed methods use the child architecture specified above. The child network is one hidden layer shallower than the parent to reduce per-subdomain cost after parent-to-child warm start while preserving the 256-dimensional Fourier embedding. This architectural convention, together with the total number of training epochs, initial collocation budget, optimizer, and loss weights, is shared by {TSDD-PINN} and the baselines, as described in Section~\ref{sec:baselines}. The full settings are reported in Table~\ref{tab:hyperparams}. Therefore, the chosen layer count does not confer a method-specific architectural advantage on {TSDD-PINN}.

The global solution is defined in a piecewise manner:
\begin{equation}\label{eq:piecewise}
\hat{u}(x,t;\theta) = \sum_{s=1}^{n} \mathbf{1}_{\Omega_s}(x,t)\;\hat{u}_s(x,t;\theta_s),
\end{equation}
where $n$ is the number of subdomains and $\mathbf{1}_{\Omega_s}$ is the indicator function. Each point in the domain is handled by exactly one sub-network, permitting sharp transitions at subdomain interfaces when physically warranted (e.g., at shockwave locations).

\subsection{Loss Function}\label{sec:loss}

The training objective is a weighted sum of three terms:
\begin{equation}\label{eq:total-loss}
\mathcal{L}_{\mathrm{total}}(\theta) = w_{\mathrm{data}}\,\mathcal{L}_{\mathrm{data}}(\theta) + w_{\mathrm{pde}}\,\mathcal{L}_{\mathrm{pde}}(\theta) + w_{\mathrm{int}}\,\mathcal{L}_{\mathrm{int}}(\theta),
\end{equation}
with fixed weights $w_{\mathrm{data}} = 0.85$, $w_{\mathrm{pde}} = 0.05$, and $w_{\mathrm{int}} = 0.10$, which are held constant across datasets, baselines, and seeds.

The \textbf{data loss} measures the mean squared error between predictions and sensor observations:
\begin{equation}\label{eq:data-loss}
\mathcal{L}_{\mathrm{data}}(\theta) = \frac{1}{N_{\mathrm{data}}} \sum_{i=1}^{N_{\mathrm{data}}} \bigl(\hat{u}(x_i^d, t_i^d;\theta) - u_i^{\mathrm{obs}}\bigr)^2.
\end{equation}

The \textbf{PDE loss} penalizes violations of the LWR equation at collocation points sampled throughout $\Omega$. When multiple subdomains are present, the PDE loss is computed in each subdomain and averaged:
\begin{equation}\label{eq:pde-loss}
\mathcal{L}_{\mathrm{pde}}(\theta) = \frac{1}{n}\sum_{s=1}^{n}\; \frac{1}{N_{\mathrm{pde}}^s} \sum_{j=1}^{N_{\mathrm{pde}}^s} w_j^{\mathrm{causal}}\;\bigl(r(x_j^c, t_j^c;\theta_s)\bigr)^2,
\end{equation}
where $r(\cdot)$ is the normalized residual from Equation~\eqref{eq:residual}. The causal weights $w_j^{\mathrm{causal}}$ are constant within each temporal bin and are defined in Section~\ref{sec:additional-components}. The \textbf{interface loss} $\mathcal{L}_{\mathrm{int}}$ enforces consistency between adjacent subdomains and is defined in Section~\ref{sec:interface}. Prior to decomposition (Stage~1), $\mathcal{L}_{\mathrm{int}} = 0$ since only a single subdomain exists.

The weight $w_{\mathrm{data}} = 0.85$ reflects the dominance of data fidelity in this data assimilation setting: sensor observations provide direct evidence of the traffic state, while the PDE serves as a regularizer that propagates information to unobserved regions. The relatively small $w_{\mathrm{pde}} = 0.05$ prevents the physics loss from overwhelming the data loss, consistent with the findings of \citet{lei2025potential} that excessive physics weighting can cause PINNs to fail for traffic flow estimation.

\subsection{Two-Stage Training Framework}\label{sec:two-stage}

{TSDD-PINN} employs a two-stage training procedure that separates coarse global learning (Stage~1) from fine-grained local refinement (Stage~2). Between the stages, the controlled path fixes refinement on, while the operational path applies the prespecified screen; whenever refinement is active, residual analysis places the spatial split. This observation-aligned two-stage domain-decomposition structure is the methodological contribution.

\paragraph{Stage 1: Coarse PINN (epochs 1 to $E_{\mathrm{split}}$)} A single neural network is trained on the full domain $\Omega$, identically to a standard (vanilla) PINN except that causal weighting (Section~\ref{sec:additional-components}) is applied to the PDE loss. The Adam optimizer \citep{kingma2014adam} is used with a learning rate of $10^{-3}$. During Stage~1, domain decomposition, residual-adaptive refinement (RAR), and interface conditions are all disabled. Stage~1 is used to produce
a global parent and a residual profile that indicates where this coarse approximation has difficulty. The residual may reflect physical gradients, optimization, sampling, or loss balance. In our experiments, $E_{\mathrm{split}} = 5{,}000$ epochs.

\paragraph{Decomposition step: Operational screening and residual-based partitioning} At epoch $E_{\mathrm{split}}$, the coarse PINN solution is analyzed to determine whether and where the domain should be partitioned. For the screen-governed operational path, a prespecified data-gradient ratio determines whether the screen activates. It is neither a validated shock detector nor a predictor of beneficial refinement. Given sensor observations $\mathcal{O}=\{(x_i,t_k,u_{i,k})\}$, we compute per-pair mean spatial gradients $g^{x}_{i}$ over timestamps common to adjacent sensors $i$ and $i{+}1$, and per-sensor mean temporal gradients $g^{t}_{i}$ from consecutive samples at sensor $i$. The shock indicator is the dimensionless max-to-mean ratio of these gradients,
\begin{equation}\label{eq:shock-indicator}
\mathcal{S}(\mathcal{O})\;=\;\max\!\left\{\frac{\max_i g^{x}_{i}}{\overline{g^{x}}+\varepsilon},\;\frac{\max_i g^{t}_{i}}{\overline{g^{t}}+\varepsilon}\right\},\qquad \varepsilon=10^{-10},
\end{equation}
and, only on the screen-governed operational path, the screen activates when $\mathcal{S}(\mathcal{O})>\tau_{\mathrm{shock}}$, with $\tau_{\mathrm{shock}}=2.0$ (Table~\ref{tab:hyperparams}). The ratio is invariant to global rescaling of $u$, but the threshold is a fixed protocol choice rather than an optimized or validated selector. The controlled path fixes refinement on regardless of $\mathcal{S}$. Whenever refinement is active, the coarse residual profile determines split positions through Section~\ref{sec:adaptive-detection}.
Within the TSDD-PINN framework, the decomposition-enabled controlled evaluation specifies refinement in advance. Under the prespecified baseline path, the screen is applied first, and training continues on a single domain when the screen does not trigger. The optional no-trigger rule returns the captured Stage~1 estimate before any additional refinement or training. Prespecified implementation checks confirmed exact Stage~1 return in two no-trigger cases and unchanged refinement in one trigger case. The controlled evaluation is unaffected because refinement is specified in advance. The no-trigger rule does not validate the screen.
When decomposition is used, child networks are created and initialized from the parent weights as described in Section~\ref{sec:adaptive-detection}. The behavior on a no-trigger operational path depends on whether the original continuation or the optional early exit is selected.

\paragraph{Stage 2: Fine-tuning after the transition step (epochs $E_{\mathrm{split}}+1$ to $E_{\max}$)} After decomposition, all subdomain networks are trained simultaneously using the following Stage~2 settings and refinement-specific operations:

\begin{itemize}
\item \emph{Reduced learning rate}: Adam with $\mathrm{lr} = 10^{-4}$ and a step decay schedule (factor $0.9$ every $5{,}000$ epochs).
\item \emph{Interface conditions}: Smooth interface continuity is applied in the evaluated refinement procedure. Rankine--Hugoniot and entropy penalties are available via a conditional branch that did not activate during the reported component analysis (Section~\ref{sec:interface}).
\item \emph{Residual-adaptive refinement}: Collocation points are periodically added in high-residual regions (Section~\ref{sec:additional-components}).
\item \emph{Causal weighting}: The temporal weighting used in Stage~1 is retained in the Stage~2 PDE loss.
\item \emph{Gradient clipping}: Gradient norms are clipped to a maximum of $5.0$ to prevent training instabilities after decomposition.
\end{itemize}

In our experiments, $E_{\max} = 20{,}000$ epochs, so Stage~2 runs for $15{,}000$ epochs. No further domain splits occur during Stage~2. \ref{app:algorithm} provides the full training pseudocode.

\subsection{Operational Screening and Residual-Based Partitioning}\label{sec:adaptive-detection}

At epoch $E_{\mathrm{split}}$, {TSDD-PINN} analyzes the spatial distribution of the PDE residual to compute candidate split positions for the decomposition-enabled stage. On the screen-governed operational path, this residual-guided split placement is preceded by the fallback decision in Section~\ref{sec:two-stage}. Thus, the method reduces manual specification of split locations when decomposition is used, while a no-trigger decision returns the captured Stage~1 predictor; that return does not validate the screen.

\begin{enumerate}
\item \textbf{Residual evaluation.} The normalized PDE residual $r(x,t)$ from Equation~\eqref{eq:residual} is evaluated on a uniform grid of $n_x \times n_t$ points ($n_x = 200$, $n_t = 100$) spanning the full domain.

\item \textbf{Spatial residual profile.} The squared residual is averaged over the temporal dimension to obtain a one-dimensional spatial profile:
\begin{equation}\label{eq:residual-profile}
R(x) = \frac{1}{n_t}\sum_{j=1}^{n_t} r(x, t_j)^2.
\end{equation}
Peaks in $R(x)$ indicate spatial locations where the coarse PINN has the highest PDE error. They are measures of model difficulty rather than validated shock locations.

\item \textbf{Smoothing.} A moving average filter with kernel size $K = \max(3, \lfloor n_x/20 \rfloor)$ (enforced to be odd) is applied to $R(x)$ to suppress noise and ensure robust peak detection.

\item \textbf{Peak detection.} Local maxima of the smoothed profile that exceed $30\%$ of the global maximum are identified, subject to a minimum inter-peak distance of $10\%$ of $n_x$ grid points. The outermost $10\%$ of the domain on each side is excluded from the peak search to avoid boundary artifacts. The number of detected peaks determines the number of subdomains: $k$ peaks yield $k+1$ candidate subdomains.

\item \textbf{Valley detection and split positioning.} All local minima in the smoothed profile are identified. The $k$ deepest minima (those with the smallest $R(x)$ values) are selected as split positions. This is a deterministic partition heuristic; it is not an optimal-interface rule or a general separator of congested and uncongested regions.
This heuristic is most effective when shockwave positions are relatively stationary over the observation window; for rapidly moving shocks, temporal averaging of the residual may reduce localization precision. If fewer minima exist than required, equally spaced splits are used as a fallback. Split positions are validated to maintain a minimum spacing $\delta_{x,\min}$ from domain edges and from each other.

\item \textbf{Parent-to-child initialization.} The domain is partitioned at the selected split positions, and each child subdomain $\Omega_s$ is assigned a network $\mathrm{NN}_s$. If the child architecture differs from the parent (fewer hidden layers), compatible layers are copied directly; otherwise, the full parent state is loaded. Each child network then undergoes 200 additional initialization epochs in which it is trained (via Adam, $\mathrm{lr}=10^{-3}$) to match the parent network's output at 2{,}000 random points within its subdomain. This makes the piecewise solution immediately after decomposition approximate the coarse PINN solution, providing a warm start for Stage~2.
\end{enumerate}

This procedure performs spatial decomposition based on the time-averaged residual profile $R(x)$, the default decomposition direction evaluated in Section~\ref{sec:direction}. The same residual-averaging procedure is used for the other decomposition directions in the ablation study of Section~\ref{sec:results-ablation}. For temporal decomposition, the analogous profile $R(t) = (1/n_x)\sum_i r(x_i,t)^2$ is computed by averaging the squared residual over the spatial dimension, and split positions in time are identified at residual valleys using the same peak and valley detection logic. For space-time decomposition, both $R(x)$ and $R(t)$ are computed independently, and their respective valley positions are used jointly to create a $2\times 2$ subdomain partition. Interface treatment is assigned by decomposition direction, as detailed in Section~\ref{sec:interface}.

\subsection{Interface Conditions}\label{sec:interface}

At each spatial interface $x_{\mathrm{int}}$ between adjacent subdomains $\Omega_L$ and $\Omega_R$, {TSDD-PINN} applies smooth interface coupling with optional conditional conservation terms. Let $\hat{u}_L(t) = \hat{u}_L(x_{\mathrm{int}}^-, t)$ and $\hat{u}_R(t) = \hat{u}_R(x_{\mathrm{int}}^+, t)$ denote the left and right predictions at the interface, evaluated at $N_{\mathrm{int}} = 200$ random time samples per training step.

Each interface is classified as either \emph{shock} or \emph{smooth} based on the mean density jump. Let $\hat{\rho}_L$ and $\hat{\rho}_R$ be the densities obtained from the predicted speeds via $\hat{\rho} = \hat{\rho}_{\mathrm{jam}}(1 - \hat{u}/\hat{u}_{\max})$, where $\hat{\rho}_{\mathrm{jam}}$ and $\hat{u}_{\max}$ both equal $1$ in the normalized coordinate system. The corresponding normalized flux is $\hat{q}(\hat{\rho})=\hat{u}_{\max}\hat{\rho}(1-\hat{\rho}/\hat{\rho}_{\mathrm{jam}})=\hat{\rho}(1-\hat{\rho})$. If the mean absolute density jump $\overline{|\hat{\rho}_L - \hat{\rho}_R|}$ exceeds a threshold $\delta_{\mathrm{shock}}$, the interface is classified as a shock; otherwise, it is treated as smooth.

\paragraph{Smooth interface: $C^0$/$C^1$ continuity} When no shock is present, the solution should be smooth across the interface. The loss enforces both value and gradient continuity:
\begin{equation}\label{eq:smooth-interface}
\mathcal{L}_{\mathrm{smooth}} = \frac{1}{N_{\mathrm{int}}}\sum_{m=1}^{N_{\mathrm{int}}} \bigl(\hat{u}_L(t_m) - \hat{u}_R(t_m)\bigr)^2 + \frac{1}{N_{\mathrm{int}}}\sum_{m=1}^{N_{\mathrm{int}}} \bigl(\partial_x \hat{u}_L(t_m) - \partial_x \hat{u}_R(t_m)\bigr)^2.
\end{equation}

Before specifying the interface loss, we clarify the intended role of these fixed interfaces. \allowbreak{TSDD-PINN}'s subdomain boundaries are artificial fixed spatial interfaces, not parameterizations of the true moving shock trajectory. The optional Rankine-Hugoniot and entropy terms introduced below are conditional conservation terms that discourage nonphysical mass imbalance across these fixed artificial interfaces; they do not localize shocks or exactly track trajectories. An explicit approach is given by \citet{lorin2024nondiffusive}. Applications requiring exact shock geometry are better served by weak-form or explicit discontinuity-line formulations \citep{deryck2024wpinns,lorin2024nondiffusive,neelan2025physics}; the present formulation trades that rigor for simplicity of integration with sparse fixed-sensor data and automatic differentiation.

\paragraph{Shock interface: Rankine-Hugoniot and entropy conditions} When a shock is detected, enforcing $C^0$/$C^1$ continuity would be physically incorrect because the LWR solution is discontinuous at the shock. Instead, we impose the Rankine-Hugoniot (R-H) condition \citep{rankine1870xv, hugoniot1887memoire}, which ensures conservation of vehicles across the discontinuity. We introduce a trainable shock speed $s$ for each interface and penalize deviations from the R-H relation:
\begin{equation}\label{eq:shock}
\mathcal{L}_{\mathrm{RH}} = \frac{1}{N_{\mathrm{int}}}\sum_{m=1}^{N_{\mathrm{int}}} \Bigl[s\bigl(\hat{\rho}_L(t_m) - \hat{\rho}_R(t_m)\bigr) - \bigl(\hat{q}(\hat{\rho}_L(t_m)) - \hat{q}(\hat{\rho}_R(t_m))\bigr)\Bigr]^2.
\end{equation}
The shock speed $s$ is initialized to zero and updated via stochastic gradient descent with a learning rate of $10^{-3}$.

For the physical Greenshields flux, the characteristic speed is $\lambda(\rho)=\mathrm{d}q/\mathrm{d}\rho=v_f(1-2\rho/\rho_{\mathrm{jam}})$. Its normalized form is $\hat{\lambda}(\hat{\rho})=\hat{u}_{\max}(1-2\hat{\rho}/\hat{\rho}_{\mathrm{jam}})=1-2\hat{\rho}$. To select the physically admissible (entropy) solution among possible weak solutions, we additionally enforce the Lax entropy condition, $\hat{\lambda}(\hat{\rho}_L) \geq s \geq \hat{\lambda}(\hat{\rho}_R)$:
\begin{equation}\label{eq:entropy}
\mathcal{L}_{\mathrm{entropy}} = \frac{1}{N_{\mathrm{int}}}\sum_{m=1}^{N_{\mathrm{int}}} \Bigl[\mathrm{ReLU}\bigl(s - \hat{\lambda}(\hat{\rho}_L(t_m))\bigr)^2 + \mathrm{ReLU}\bigl(\hat{\lambda}(\hat{\rho}_R(t_m)) - s\bigr)^2\Bigr].
\end{equation}
This condition requires that characteristics propagate \emph{into} the shock from both sides. $\hat{\lambda}$ and the learned interface speed $s$ are dimensionless characteristic speeds in normalized $\hat{x}/\hat{t}$ coordinates and are not directly comparable with the physical ASM wave speeds. The total shock interface loss is $\mathcal{L}_{\mathrm{shock}} = \mathcal{L}_{\mathrm{RH}} + w_{\mathrm{entropy}}\,\mathcal{L}_{\mathrm{entropy}}$, where $w_{\mathrm{entropy}} = 1.0$.

The overall interface loss selects between the two modes based on the classified interface type:
\begin{equation}\label{eq:int-loss}
\mathcal{L}_{\mathrm{int}} = \sum_{\text{interfaces}} \begin{cases} \mathcal{L}_{\mathrm{shock}} & \text{if } \overline{|\hat{\rho}_L - \hat{\rho}_R|} > \delta_{\mathrm{shock}}, \\ \mathcal{L}_{\mathrm{smooth}} & \text{otherwise.} \end{cases}
\end{equation}
Smooth interface coupling is applied at every interface; the R-H/entropy terms form a separate conditional branch. The component analysis does not treat inactivity of the conditional branch as inactivity of all interface coupling.

The interface conditions defined above contrast with XPINN's residual-based continuity \citep{jagtap2020extended}, which does not explicitly encode the R-H conservation relation. For the temporal and space-time ablations (Section~\ref{sec:results-ablation}), spatial sub-interfaces use the same R-H treatment, whereas temporal sub-interfaces enforce $C^0$ solution continuity, $\mathcal{L}_{\mathrm{cont},t} = (1/N_{\mathrm{int}})\sum_m [\hat{u}_{\mathrm{after}}(x_m, t^\ast) - \hat{u}_{\mathrm{before}}(x_m, t^\ast)]^2$, because the R-H relation is a spatial shock condition.

\subsection{Additional Training Components}\label{sec:additional-components}

\paragraph{Residual-adaptive refinement (RAR)} During Stage~2, additional collocation points are periodically injected in high-residual regions to concentrate training effort where the current approximation has large residual \citep{wu2023comprehensive}. Every $2{,}500$ epochs, $5{,}000$ candidate points are sampled uniformly in each subdomain, the absolute PDE residual is evaluated at each candidate, and the $2{,}500$ points with the largest $|r|$ are added to the collocation set. This sampling rule constitutes one component of the evaluated method, and its isolated contribution is not established here.

\paragraph{Causal weighting} The PDE loss applies a temporal weighting scheme inspired by \citet{wang2024respecting} that assigns a higher weight to earlier time steps throughout training (both Stage~1 and Stage~2). The rationale is that PDE solutions should be solved sequentially in time: errors at early times propagate forward, so the network should prioritize accuracy at earlier times. Concretely, the collocation points are sorted by their temporal coordinate and partitioned into $n_{\mathrm{bins}}$ equal-sized bins. The causal weight for bin $j$ is
\begin{equation}\label{eq:causal}
w_j^{\mathrm{causal}} = \exp\!\Bigl(-\epsilon \sum_{k < j} \bar{r}_k^2\Bigr),
\end{equation}
where $\bar{r}_k^2$ is the mean squared residual in bin $k$ and $\epsilon$ is a causality parameter. The weights are detached from the computational graph (treated as constants during backpropagation) so that gradients flow only through the residuals, not through the weighting scheme.

\paragraph{Gradient clipping} All parameter gradients are clipped to a maximum norm of $5.0$ at every training step. This prevents training instabilities caused by large gradients that can arise immediately after domain decomposition, when the interface loss introduces new terms into the objective.

\subsection{Decomposition Direction Analysis}\label{sec:direction}

The TSDD-PINN framework supports spatial, temporal, and space--time decomposition. The controlled benchmark uses the spatial implementation, while Section~\ref{sec:results-ablation} evaluates all three directions under matched conditions. Spatial decomposition attains the lowest mean error and requires less than half the training time of space--time decomposition in the tested fixed-sensor setting, while temporal decomposition is faster. This result does not constitute a general direction-selection rule. Three considerations motivate the comparison.

\paragraph{Argument 1: Shockwave geometry} Shockwaves in LWR traffic flow manifest primarily as spatially localized transition regions: at any given time, the speed field exhibits a sharp gradient across a congestion boundary in the $x$-direction. In our fixed-sensor, lane-aggregated, finite-resolution setting, the corresponding temporal speed trace at a given sensor location typically varies more gradually than the spatial profile across the transition. This geometry motivates testing spatial refinement. It neither specifies the interface location nor establishes that spatial decomposition outperforms temporal alternatives.

\paragraph{Argument 2: Data coverage asymmetry} Fixed sensors provide fundamentally asymmetric coverage of the spatiotemporal domain. In the spatial direction, sensors occupy only a small fraction of the domain (e.g., 5 sensors covering 5 out of 100 spatial cells, or 5\% spatial coverage). In the temporal direction, each sensor records a continuous time series spanning the entire observation period (effectively 100\% temporal coverage). Spatial decomposition adds no observations and does not shorten the
physical sensor gaps. It only allocates separate model capacity across
the unobserved intervals. This asymmetry motivates, but does not prove,
the spatial-direction test.

\paragraph{Argument 3: Residual anisotropy} The coarse residual can be spatially structured, but it can reflect approximation, optimization, sampling, and loss balance rather than a physical shock alone. Thus, we use $R(x)$ only as a deterministic model-difficulty heuristic. The residual evidence does not validate localization or establish spatial error dominance; the direction ablation in Section~\ref{sec:results-ablation} supplies the empirical comparison.

We note that temporal decomposition may be beneficial in other TSE contexts, such as probe-vehicle-based estimation, where spatial coverage is dense but temporal sampling is sparse. Evaluating such settings is outside the scope of the present study.

\section{Experimental Setup}\label{sec:experiments}

\subsection{Datasets}\label{sec:datasets}

The empirical evaluation uses two trajectory datasets, each with a distinct role. The I-24 MOTION dataset serves as the primary benchmark for evaluating {TSDD-PINN} under sparse fixed-sensor reconstruction with localized congestion and shock-prone transition regions. The NGSIM I-80 dataset supplies a supplementary negative-control comparison among the existing no-trigger continuation, forced decomposition, and Stage~1/early exit; it is not a selector-validation experiment. The data collection sites are shown in Figure~\ref{fig:trajectory-datasets}.

\paragraph{I-24 MOTION} We evaluate {TSDD-PINN} on five days of the I-24 MOTION dataset \citep{gloudemans202324}, which captures lane-level vehicle trajectories along a 4.2-mile (21{,}120~ft) corridor of Interstate~24 near Nashville, Tennessee, using 276 overhead cameras at approximately 2-second resolution. The raw trajectory data are aggregated into a spatiotemporal grid of $100$ spatial cells ($\Delta x \approx 211$~ft) and $7{,}200$ time steps ($\Delta t = 2$~s), covering a 4-hour observation window from 06:00 to 10:00~CST.
The five days span a range of traffic conditions, from mild free-flow to severe incident-induced congestion, as summarized in Table~\ref{tab:datasets}. This diversity enables the primary evaluation of {TSDD-PINN} under both recurrent and non-recurrent congestion on a long freeway corridor.

\begin{table}[!htbp]
\centering
\caption{Summary of the primary I-24 MOTION datasets. All I-24 datasets span a 4-hour window from 06:00 to 10:00 CST. The congestion ratio is defined as the percentage of spatiotemporal cells in which the observed speed is below 30 mph.}
\label{tab:datasets}
\small
\begin{tabularx}{\textwidth}{@{}l c >{\raggedright\arraybackslash}X c@{}}
\toprule
Date & Day & Traffic condition & Cong.\ (\%) \\
\midrule
2022-11-21 & Mon & Rear-end collision WB near MM~59.7; congestion 06:14-07:43 & 55 \\
2022-11-22 & Tue & No incidents; typical weekday recurrent congestion & 47 \\
2022-11-23 & Wed & Sideswipe WB lane~1 at MM~59.2; congestion 07:35-07:45 & 11 \\
2022-11-29 & Tue & No incidents; severe recurrent congestion & 63 \\
2022-12-02 & Fri & Two-vehicle crash WB near MM~61.8; 07:38-08:20 & 37 \\
\bottomrule
\end{tabularx}
\end{table}

\paragraph{NGSIM I-80} We additionally use the NGSIM I-80 trajectory dataset as a supplementary negative-control case \citep{usdot_ngsim_2016,huang2022physics}. The dataset contains detailed vehicle trajectories collected from a 1{,}600-ft segment of the I-80 freeway in Emeryville, California. Following prior traffic PINN studies, we use the 15-min PM peak interval from 16:00 to 16:15 on April 13, 2005 and aggregate the speed field using a spatial resolution of $\Delta x = 20$~ft and a temporal resolution of $\Delta t = 5$~s. NGSIM is not primary evidence for decomposition accuracy. In 50 historical runs, the prespecified screen does not activate, enabling comparison of the existing continuation, forced-decomposition reference, and Stage~1/early exit. This outcome does not establish the absence of a localized boundary, a regime-adaptive mechanism, or selector quality. The same sensor counts, $n_s \in \{3,4,5,6,7\}$, ten random seeds, and relative $L_2$ error metric are used for the NGSIM supplementary experiment.

Speed values are normalized to $[0,1]$ via min-max scaling: $\hat{u} = (u - u_{\min})/(u_{\max} - u_{\min})$, where $u_{\min}$ and $u_{\max}$ are computed from the full trajectory-derived aggregated speed field in each dataset. The free-flow speed $v_f$ is estimated as the 95th percentile of the full aggregated field and held fixed throughout training. Since these statistics are derived from the full field rather than the sparse sensor subset, the present study should be read as an offline reconstruction evaluation. Online or streaming deployment would require these quantities to be estimated from sensor observations alone or from historical aggregates; the sparse-only sensitivity is summarized in Section~\ref{sec:results-sensor}, and the remaining limitation is discussed in Section~\ref{sec:disc-limitations}. Given a domain discretized into $N$ spatial cells, $n_s$ sensors are placed at equally spaced \emph{interior} positions by generating $n_s + 2$ equally spaced indices from $0$ to $N-1$ and discarding the two boundary indices. Formally, the candidate indices are $\mathrm{linspace}(0, N-1, n_s+2)$, and the $n_s$ interior elements (excluding the first and last) are retained and rounded to the nearest integer. For example, with $n_s = 5$, the sensor cell indices are approximately $\{16, 33, 49, 66, 82\}$ for I-24 MOTION ($N=100$) and $\{13, 26, 40, 53, 66\}$ for NGSIM I-80 ($N=80$). These are virtual fixed sensors sampled from trajectory-derived aggregated fields rather than physical loop-detector records. This design enables controlled comparison across sensor counts and full-field error computation, but the evidence should be interpreted as offline reconstruction rather than online deployment.

\subsection{Baseline Methods}\label{sec:baselines}

We compare {TSDD-PINN} (B6) against five baselines that represent a progression from data-driven to increasingly sophisticated physics-informed approaches. The neural and PINN-family benchmark does not imply superiority over untested classical estimators. The two non-neural comparators are canonical time-slice spatial linear interpolation and the adaptive smoothing method (ASM) \citep{treiber2002reconstructing}. For each I-24 time row, linear interpolation reconstructs the spatial field between adjacent selected sensors, uses the same 100-cell evaluation grid as the neural estimators, and holds the nearest selected-sensor value outside the sensor span. It is matched to both the observation and evaluation grids and uses all 7{,}200 time rows. ASM reconstructs free-flow and congested fields using anisotropic space--time kernels and combines them through a speed-dependent transition weight. The published characteristic wave speeds, temporal bandwidth, and speed-transition parameters are fixed. Only the spatial bandwidth is selected between two prespecified values by leave-one-sensor-out cross-validation (LOOCV): each selected sensor is withheld in turn, reconstructed from the remaining sensors, and the pooled relative $L_2$ error across the held-out traces selects the bandwidth. The selected bandwidth is then applied using all selected sensors before the complete field is evaluated, so no unobserved cell informs its selection. ASM is matched to the selected sensor locations, is evaluated on its own 360-row time grid, and retains kernel-based extrapolation outside the sensor span. Section~\ref{sec:results-sensor} reports the results, and~\ref{app:classical-details} gives the equations, parameter values, and calibration details. Table~\ref{tab:baselines} summarizes the six neural and PINN-family methods; the two non-neural comparators are defined separately above.

\begin{table}[!htbp]
\centering
\caption{Component comparison of evaluated methods.}
\label{tab:baselines}
\small
\setlength{\tabcolsep}{5pt}
\begin{tabularx}{\textwidth}{@{}>{\raggedright\arraybackslash}X *{6}{c}@{}}
\toprule
 & B1 & B2 & B3 & B4 & B5 & B6 \\
Component & NN & PINN & RAR & Visc. & XPINN & Ours \\
\midrule
PDE constraint (LWR) &  & \checkmark & \checkmark & \checkmark & \checkmark & \checkmark \\
Adaptive collocation (RAR) &  &  & \checkmark &  &  & \checkmark \\
Viscosity regularization &  &  &  & \checkmark &  &  \\
Domain decomposition &  &  &  &  & \checkmark & \checkmark \\
\multicolumn{1}{@{}l}{Data-dependent partition heuristic}&  &  &  &  &  & \checkmark \\
Two-stage training &  &  &  &  &  & \checkmark \\
Smooth coupling; optional R-H/entropy terms &  &  &  &  &  & \checkmark \\
Causal weighting &  &  &  &  &  & \checkmark \\
Parent weight initialization &  &  &  &  &  & \checkmark \\
\bottomrule
\end{tabularx}
\par\vspace{4pt}
{\small References: B2 \cite{raissi2019physics,huang2022physics}; B3 \cite{wu2023comprehensive}; B4 \cite{huang2022physics,huang2023limitations}; B5 \cite{jagtap2020extended}; B6 ({TSDD-PINN}, this work).}
\end{table}

\begin{itemize}
\item \textbf{B1 (Neural network).} A feedforward network with the same architecture as {TSDD-PINN}'s parent, trained on sensor data only without any PDE constraint.
\item \textbf{B2 (Vanilla PINN)} \citep{raissi2019physics, huang2022physics}. A standard PINN minimizing the weighted sum of data and PDE losses with fixed weights ($w_{\mathrm{data}} = 0.85$, $w_{\mathrm{pde}} = 0.05$). No decomposition or adaptive refinement.
\item \textbf{B3 (PINN + RAR)} \citep{wu2023comprehensive}. A vanilla PINN augmented with residual-adaptive refinement: every 2{,}500 epochs, 2{,}500 new collocation points are added in regions of high PDE residual. Isolates the effect of adaptive collocation from decomposition.
\item \textbf{B4 (PINN + Viscosity)} \citep{huang2022physics, huang2023limitations}. A PINN that adds an artificial viscosity term $\varepsilon \, \partial^2 u / \partial x^2$ (with $\varepsilon = 0.1$) to the LWR PDE, motivated by the finding that standard PINNs cannot accurately represent non-smooth LWR solutions \citep{huang2023limitations}.
\item \textbf{B5 (XPINN)} \citep{jagtap2020extended}. The extended PINN framework with $n_x \times n_t = 2 \times 2 = 4$ space-time subdomains at equally spaced splits, each with the same child architecture as {TSDD-PINN} ($[2, 256, 128, 128, 1]$). Interface coupling follows \citet{jagtap2020extended} via residual continuity plus a solution-average penalty. Networks are initialized randomly without parent transfer and trained jointly from epoch~0 for the full $20{,}000$ epochs. To ensure a fair comparison, XPINN uses the same loss weights as {TSDD-PINN} ($w_{\mathrm{data}} = 0.85$, $w_{\mathrm{pde}} = 0.05$, $w_{\mathrm{int}} = 0.10$), the same StepLR schedule, and the same gradient clipping. XPINN is used here as a standard fixed space-time DD-PINN baseline rather than as an optimized XPINN variant. The comparison evaluates {TSDD-PINN} as a coupled traffic-specific design, not the isolated effect of each individual component.
\end{itemize}

All methods follow the architecture convention specified in Section~\ref{sec:network-arch}: single-domain methods use the parent architecture, while decomposed methods use the child architecture. They also share the same total number of training epochs, initial collocation budget, optimizer (Adam, initial learning rate of $10^{-3}$), and loss weights. Method-specific enhancements (Stage~2 learning-rate reduction for {TSDD-PINN}, StepLR for DD methods, RAR for B3 and B6) differ across methods.

\subsection{Evaluation and Implementation}\label{sec:eval-protocol}

Each method is evaluated on all 5 I-24 datasets, each with 5 sensor configurations ($n_s \in \{3, 4, 5, 6, 7\}$) and 10 random seeds, yielding $5 \times 5 \times 10 = 250$ runs per method and $6 \times 250 = 1{,}500$ runs in total. Sensors are placed at equally spaced interior positions as described in Section~\ref{sec:datasets}.

The primary evaluation metric is the relative $L_2$ error (\%), computed in physical speed units (mph) over all spatiotemporal grid points:
\begin{equation}\label{eq:l2-error}
\text{Relative } L_2 \text{ Error} = \frac{\sqrt{\sum_{j} \bigl(u^{\mathrm{pred}}_{\mathrm{mph}}(x_j, t_j) - u^{\mathrm{true}}_{\mathrm{mph}}(x_j, t_j)\bigr)^2}}{\sqrt{\sum_{j} \bigl(u^{\mathrm{true}}_{\mathrm{mph}}(x_j, t_j)\bigr)^2}} \times 100\%.
\end{equation}

Against each of five neural network and PINN baselines, the 25 paired units are ten-seed configuration means; lower error defines a win. For these five directional comparisons, the paired $t$-tests and Wilcoxon signed-rank tests are one-sided in the prespecified direction of lower TSDD-PINN error. The $p_{\mathrm{Holm}}$ column applies Holm--Bonferroni adjustment across the five paired $t$-tests.  Duplicate intermediate records were removed before aggregation.

\begin{table}[!htbp]
\centering
\small
\caption{Hyperparameters for I-24 MOTION experiments.}
\label{tab:hyperparams}
\begin{tabularx}{\textwidth}{@{}>{\raggedright\arraybackslash}p{0.38\textwidth} >{\raggedright\arraybackslash}X@{}}
\toprule
Parameter & Value \\
\midrule
\multicolumn{2}{l}{\emph{Architecture}} \\
Parent / Child network & $[2,256,128,128,128,1]$ / $[2,256,128,128,1]$ \\
Fourier embedding ($d_e$, $\sigma$) & 128, 10.0; activation: $\tanh$ \\
\midrule
\multicolumn{2}{l}{\emph{Training}} \\
Seeds & 42, 123, 456, 789, 1024, 2048, 3000, 4096, 5555, 7777 \\
Epochs (total / Stage 1) & 20{,}000 / 5{,}000 \\
Learning rate (Stage 1 / Stage 2) & $10^{-3}$ / $10^{-4}$ (Adam); StepLR($5000$, $0.9$) \\
Batch size (data / collocation) & 4{,}096 / $\max(512, 2048/n)$ per subdomain \\
Collocation points ($N_{\mathrm{pde}}$) & 50{,}000 (LHS) \\
Gradient clipping & max norm $= 5.0$ \\
$w_{\mathrm{data}}$, $w_{\mathrm{pde}}$, $w_{\mathrm{int}}$ (fixed) & 0.85, 0.05, 0.10 \\
\midrule
\multicolumn{2}{l}{\emph{Domain decomposition}} \\
Residual grid ($n_x \times n_t$) & $200 \times 100$ \\
Smoothing kernel & $\max(3, \lfloor n_x/20 \rfloor)$, odd \\
Peak threshold / min distance & 30\% of $\max R(x)$ / 10\% of $n_x$ \\
Min subdomain width ($\delta_{x,\min}$) & 0.15 (normalized) \\
Shock indicator & 2.0 \\
Child init (epochs / points) & 200 / 2{,}000 \\
\midrule
\multicolumn{2}{l}{\emph{RAR (Stage 2 only)}} \\
Frequency / candidates / added & Every 2{,}500 epochs / 5{,}000 / 2{,}500 \\
\midrule
\multicolumn{2}{l}{\emph{Causal weighting}} \\
Temporal bins / $\epsilon$ & 10 / 1.0 \\
\midrule
\multicolumn{2}{l}{\emph{Interface conditions}} \\
Points / $\delta_{\mathrm{shock}}$ / $w_{\mathrm{entropy}}$ & 200 / 0.1 / 1.0 \\
Shock speed learning rate & $10^{-3}$ (SGD) \\
\bottomrule
\end{tabularx}
\end{table}

All models are implemented in PyTorch and trained on a single NVIDIA RTX~3060~Ti GPU. Table~\ref{tab:hyperparams} summarizes the hyperparameters for the I-24 MOTION experiments. For reproducibility, all random seeds are fixed at the beginning of each run, controlling the random number generators in PyTorch, NumPy, and CUDA. The 10 seeds listed in Table~\ref{tab:hyperparams} are used consistently across all methods, datasets, and sensor configurations. Training time per run is approximately 500 seconds for TSDD-PINN on the I-24 datasets. Reported training times correspond to wall-clock training time (forward/backward passes, RAR sampling, interface loss, and optimizer steps), averaged across 10 seeds per configuration; data preprocessing, evaluation, and visualization are excluded. All methods are run on the same NVIDIA RTX 3060 Ti GPU under the method-specific batching protocols reported in Tables~\ref{tab:hyperparams} and~\ref{tab:efficiency}.

\section{Results}\label{sec:results}

This section evaluates {TSDD-PINN} on five I-24 MOTION days spanning accident, normal, mild, severe, and secondary accident conditions. The primary metric is the relative $L_2$ error, computed in physical speed units, over the entire space-time field.

\subsection{Main Comparison on I-24 MOTION}\label{sec:results-main}

Table~\ref{tab:main-results} reports the controlled I-24 results under the decomposition-enabled mode defined in Section~\ref{sec:two-stage}, in which decomposition is fixed across all I-24 configurations while split positions are selected from the residual profile. The table reports the mean relative $L_2$ error for all 25 configurations, each averaged over 10 random seeds. {TSDD-PINN} (B6) achieves the lowest error in 18 of 25 configurations, while XPINN (B5) is best in 6 and PINN+RAR (B3) is best in 1. The advantage of {TSDD-PINN} is concentrated in the sparse regime. For $n_s \in \{3,4,5\}$, {TSDD-PINN} ranks first in 14 of 15 configurations.

\begin{table}[!htbp]
\centering
\caption{Decomposition-enabled evaluation: mean relative $L_2$ error (\%) on I-24 MOTION datasets. Each entry is the mean over 10 random seeds. The lowest error in each row is shown in \textbf{bold}. B1: NN, B2: Vanilla PINN, B3: PINN+RAR, B4: PINN+Viscosity, B5: XPINN, B6: {TSDD-PINN} (Ours).}
\label{tab:main-results}
\small
\begin{tabular}{ll rrrrrr}
\toprule
Date & $n_s$ & B1 & B2 & B3 & B4 & B5 & B6 \\
\midrule
\multirow{5}{*}{\rotatebox{90}{20221121}}
& 3 & 43.99 & 22.66 & 21.78 & 21.59 & 21.67 & \textbf{20.69} \\
& 4 & 37.28 & 17.86 & 17.03 & 16.75 & 16.76 & \textbf{15.84} \\
& 5 & 34.34 & 16.91 & 16.42 & 16.22 & 16.04 & \textbf{15.06} \\
& 6 & 32.21 & 16.24 & 16.01 & 15.71 & 15.57 & \textbf{15.09} \\
& 7 & 26.79 & 15.62 & 14.47 & 14.26 & 14.17 & \textbf{14.07} \\
\midrule
\multirow{5}{*}{\rotatebox{90}{20221122}}
& 3 & 42.26 & 22.36 & 22.23 & 22.10 & 22.01 & \textbf{19.70} \\
& 4 & 38.30 & 20.08 & 19.46 & 19.54 & 19.26 & \textbf{17.92} \\
& 5 & 34.10 & 18.82 & 18.08 & 18.03 & 17.91 & \textbf{17.17} \\
& 6 & 31.13 & 17.10 & 16.43 & 16.32 & \textbf{16.23} & 16.48 \\
& 7 & 27.96 & 15.68 & 15.04 & 14.88 & \textbf{14.60} & 15.41 \\
\midrule
\multirow{5}{*}{\rotatebox{90}{20221123}}
& 3 & 37.50 & 11.91 & 11.22 & 11.22 & 11.15 & \textbf{10.66} \\
& 4 & 30.52 & 10.74 & 10.30 & 10.28 & \textbf{10.15} & 10.19 \\
& 5 & 24.17 & 10.25 & 9.63 & 9.63 & 9.66 & \textbf{9.61} \\
& 6 & 23.07 & 9.96 & \textbf{9.39} & 9.48 & 9.43 & 9.58 \\
& 7 & 20.81 & 9.07 & 8.79 & 8.75 & \textbf{8.65} & 8.92 \\
\midrule
\multirow{5}{*}{\rotatebox{90}{20221129}}
& 3 & 50.62 & 27.06 & 26.86 & 26.62 & 26.88 & \textbf{24.87} \\
& 4 & 47.31 & 25.79 & 25.12 & 25.15 & 25.09 & \textbf{23.54} \\
& 5 & 42.76 & 23.53 & 22.83 & 22.79 & 22.58 & \textbf{21.07} \\
& 6 & 39.02 & 22.02 & 21.22 & 21.15 & 20.97 & \textbf{20.38} \\
& 7 & 34.12 & 20.25 & 19.26 & 19.26 & \textbf{18.95} & 19.55 \\
\midrule
\multirow{5}{*}{\rotatebox{90}{20221202}}
& 3 & 40.85 & 21.04 & 20.92 & 20.64 & 20.45 & \textbf{17.98} \\
& 4 & 38.76 & 19.02 & 18.56 & 18.55 & 18.35 & \textbf{16.84} \\
& 5 & 34.22 & 18.06 & 17.48 & 17.45 & 17.06 & \textbf{16.03} \\
& 6 & 29.90 & 16.70 & 16.01 & 15.93 & 15.66 & \textbf{15.30} \\
& 7 & 25.71 & 15.13 & 14.57 & 14.36 & \textbf{14.08} & 14.65 \\
\bottomrule
\end{tabular}
\end{table}

Pairwise statistical tests across all 25 configuration means support the best-count summaries. Relative to XPINN, {TSDD-PINN} reduces the mean relative $L_2$ error by 0.67 percentage points with a 95\% confidence interval of [0.29, 1.04]. Relative to Vanilla PINN, the mean reduction is 1.49 percentage points with a 95\% confidence interval of [1.15, 1.83]. All five neural/PINN comparisons remain significant after Holm--Bonferroni correction and are supported by the Wilcoxon signed-rank tests reported in \ref{app:statistical}.

Averaged over sensor counts, {TSDD-PINN} has the lowest day-level mean on all five I-24 days. Its largest average gain occurs on 20221129, the severe day with 63\% congestion, where {TSDD-PINN} records 21.88\% relative $L_2$ error and improves on Vanilla PINN and XPINN by 1.85 and 1.01 percentage points, respectively. The smallest gain occurs on 20221123, the mild day with 11\% congestion, where {TSDD-PINN} records 9.79\% and is effectively tied with XPINN, with a 2W/3L configuration record. The accident days 20221121 and 20221202 exhibit similarly strong gains, with average improvements of 1.71 to 1.83 percentage points over Vanilla PINN and 0.69 to 0.96 percentage points over XPINN.

\begin{figure*}[!htbp]
\centering
\includegraphics[width=\textwidth]{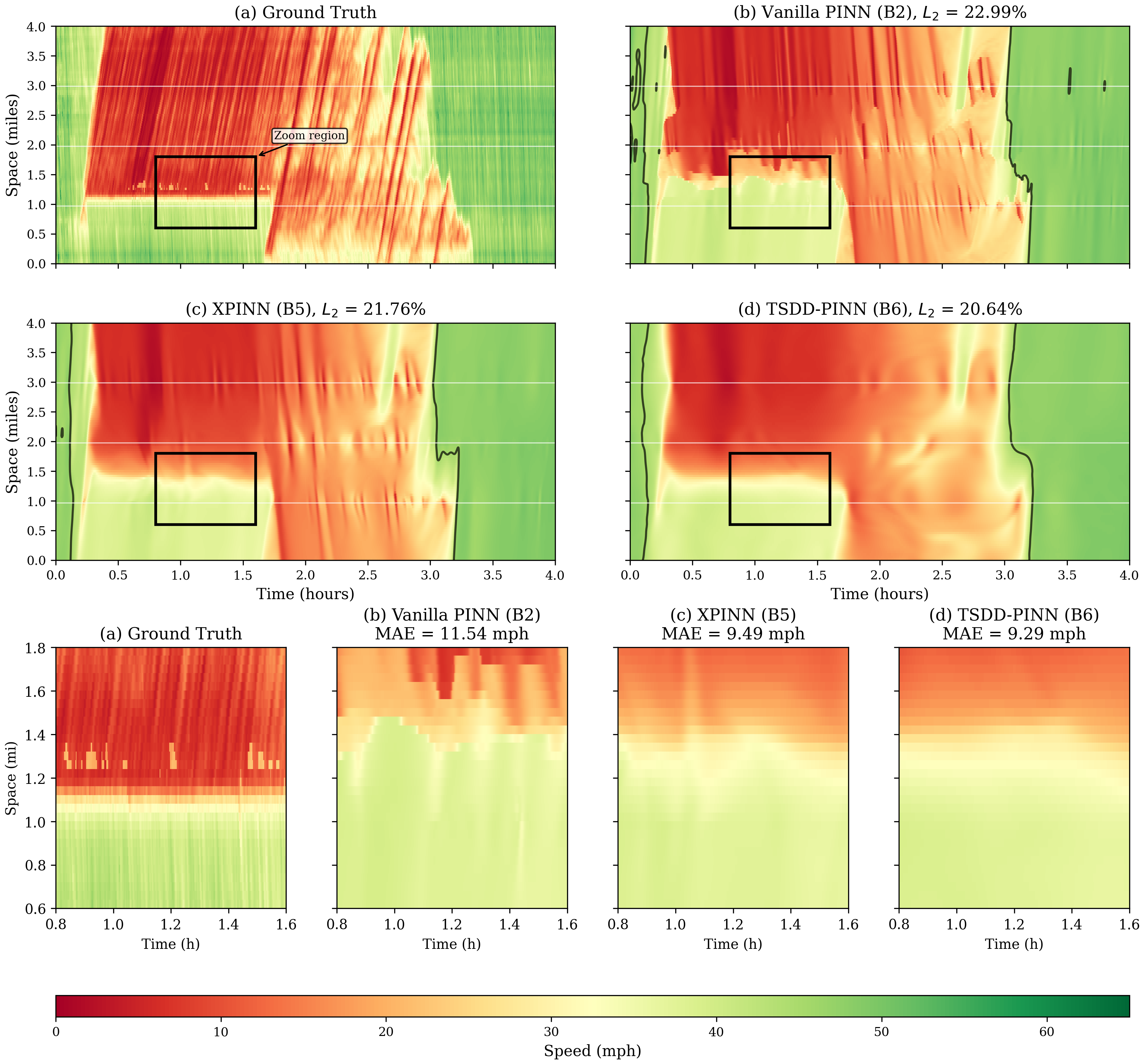}
\caption{Representative speed field reconstruction for 20221121 with 3 sensors. Panel (a) shows the ground truth. Panels (b), (c), and (d) show Vanilla PINN (B2), XPINN (B5), and {TSDD-PINN} (B6), respectively. The black box marks the zoom region spanning 0.8 to 1.6 h and 0.6 to 1.8 mi, which contains the congestion core. The 45 mph contour is overlaid on panels (b) to (d). All neural estimators recover the large-scale structure of the speed field, but the continuous function approximation smooths the sharpest gradients relative to the ground truth. In the zoom region, the mean absolute error is 11.54 mph for B2, 9.49 mph for B5, and 9.29 mph for B6; these values describe this illustrative run, not the main-table configuration means.}
\label{fig:speed-fields}
\end{figure*}

\begin{figure*}[!htbp]
\centering
\includegraphics[width=0.98\textwidth]{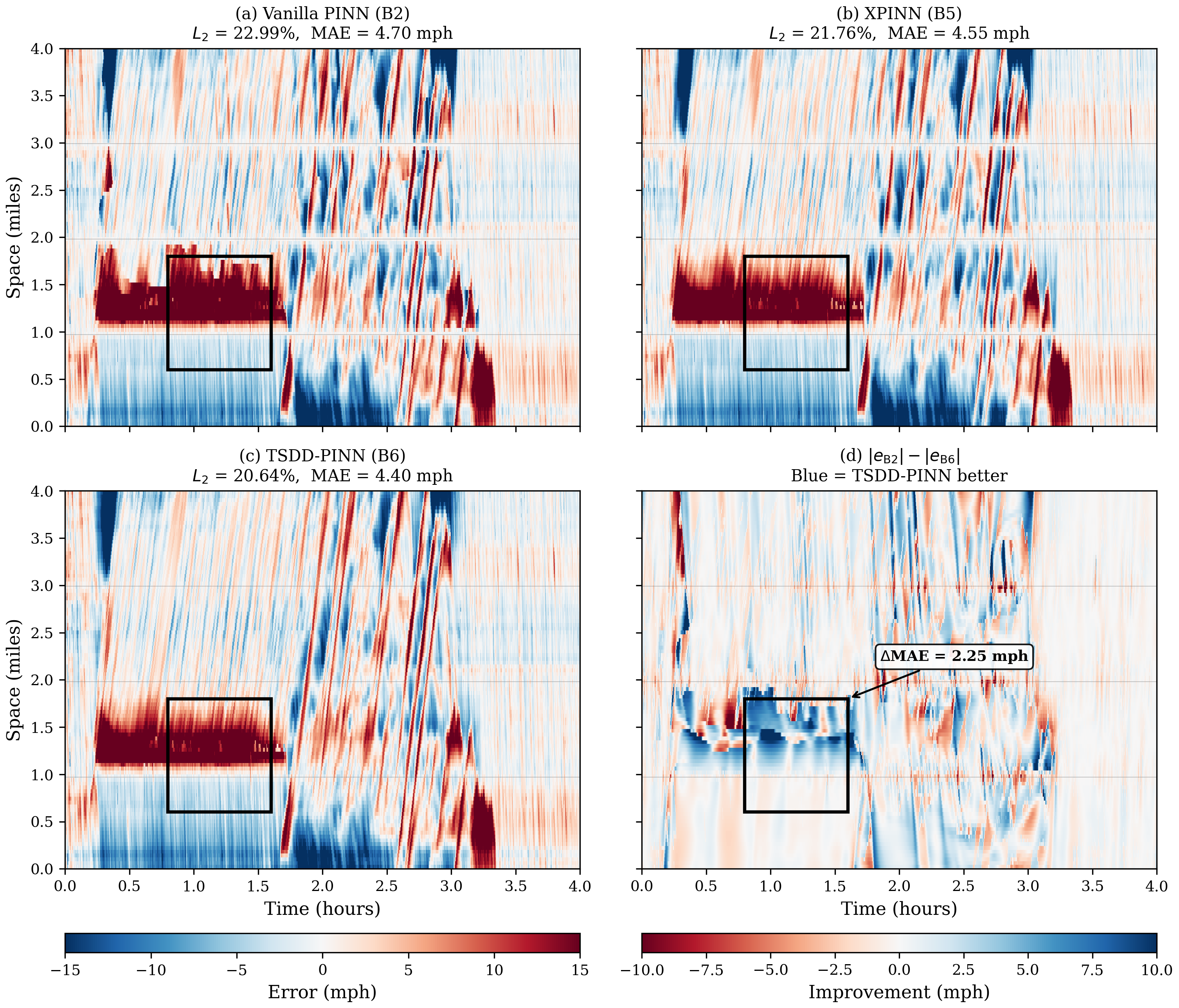}
\caption{Pointwise error maps for the representative case in Figure~\ref{fig:speed-fields}. Panels (a), (b), and (c) report the signed speed error for Vanilla PINN (B2), XPINN (B5), and {TSDD-PINN} (B6), respectively. Panel (d) shows the improvement map $|e_{\mathrm{B2}}|-|e_{\mathrm{B6}}|$, where blue indicates that {TSDD-PINN} yields lower absolute error than Vanilla PINN. The full field mean absolute error is 4.70 mph for B2, 4.55 mph for B5, and 4.40 mph for B6. In the zoom region, the mean absolute improvement of {TSDD-PINN} over Vanilla PINN is 2.25 mph. These values are descriptive and are not the main-table means.}
\label{fig:error-maps}
\end{figure*}

Figure~\ref{fig:speed-fields} presents a representative sparse case from 20221121 with three sensors, and Figure~\ref{fig:error-maps} maps the corresponding pointwise errors. Differences among the full speed fields are subtle, since all three PINN-based models recover the dominant corridor scale pattern. The visible distinction is confined to the boxed congestion core, where TSDD-PINN retains more of the low speed region than Vanilla PINN and marginally more than XPINN. The error maps resolve this difference spatially. Within the congested zone below 30 mph, which covers 54.8\% of the domain in this run, mean absolute error is 4.98 mph for TSDD-PINN and 5.55 mph for Vanilla PINN, a difference of 0.57 mph, while the methods are broadly comparable in the transition zone between 30 and 55 mph. Panel (d) shows that the regions of reduced error are concentrated along the congestion boundary and within the core rather than distributed uniformly across the domain. These observations describe a single run and are reported for illustration. They are not configuration-level evidence, and they do not isolate the contribution of any individual component.

\begin{figure}[!htbp]
\centering
\includegraphics[width=0.98\columnwidth]{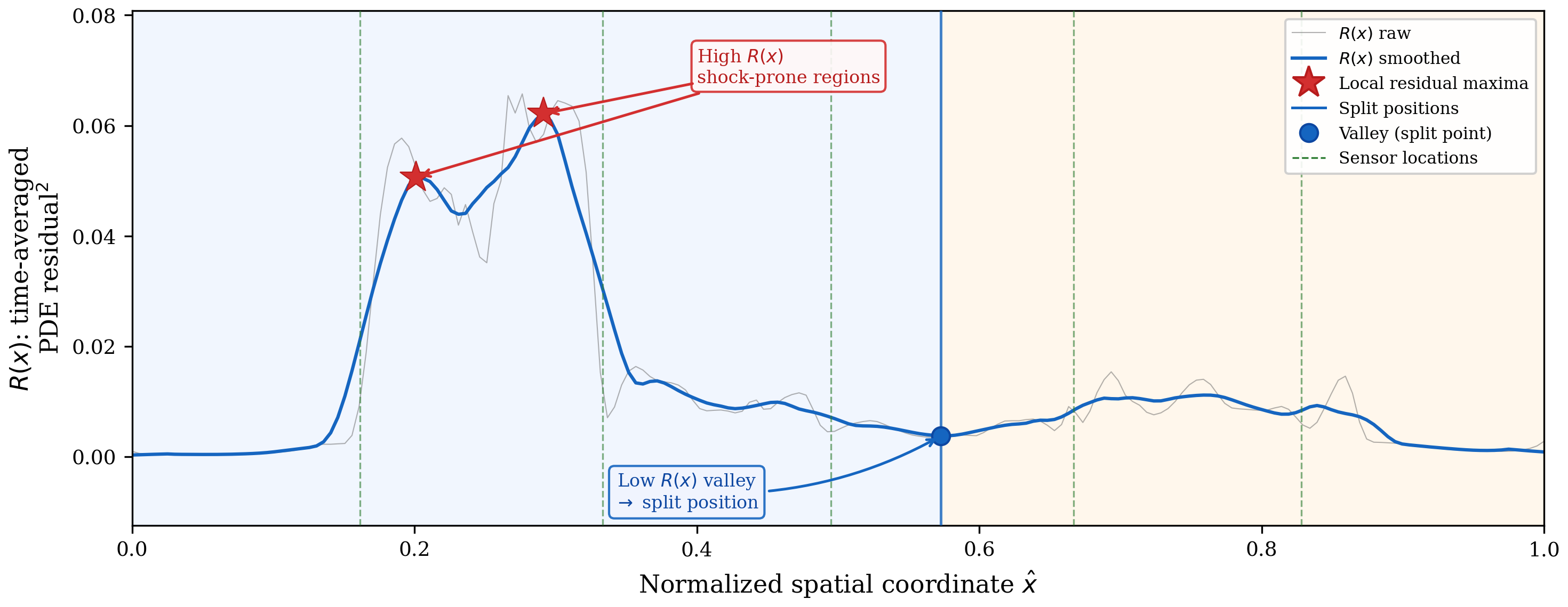}
\caption{Illustrative measure of model difficulty for 20221121 with
$n_s=5$ and seed 42. The residual maximum occurs at $x=0.3100$, the
peak of the true gradient at $x=0.2828$, the projected split at
$x=0.572864$, and the deployed valley at $x=0.577889$. These are
distinct quantities, and they do not validate shock detection,
congestion separation, or optimal interface placement.}
\label{fig:residual-profile}
\end{figure}

Figure~\ref{fig:residual-profile} serves as an illustrative measure of model difficulty. Since residual magnitude can reflect physics, optimization, sampling, or loss imbalance, the profile is used only as a deterministic partition heuristic. Localization is not better than the midpoint or random controls and fails on NGSIM. In a six-case check comparing the shares of cells below 45~mph on the two sides of the split, only two cases show separation, and only the 20221123 result is robust to partition relocation. These findings do not validate shock detection, congestion separation, or optimal interface placement.

In summary, the controlled table and descriptive figures (Table~\ref{tab:main-results} and Figures~\ref{fig:speed-fields} to \ref{fig:residual-profile}) support an overall advantage concentrated under sparse fixed sensing. They do not isolate the residual rule as its cause, and the smaller dense-sensing gains motivate the classical-comparator results in Section~\ref{sec:results-sensor}.

\subsection{Decomposition Direction Ablation}\label{sec:results-ablation}

\noindent We compare spatial, temporal, and space-time variants on two I-24 days, three sensor counts ($n_s\in\{3,5,7\}$), and ten seeds. The comparison spans 180 evaluated direction-specific results: 120 newly trained temporal and space-time runs and 60 spatial runs reused from the corresponding controlled benchmark cells.

For a representative case (20221121 with 5 sensors), spatial decomposition achieves the lowest relative $L_2$ error at 15.06\%, temporal at 16.21\%, and space-time at 15.25\% (mean training time of 583 s, 329 s, and 931 s, respectively, averaged over 10 seeds). Temporal decomposition produces the highest error in five of six configurations. The exception is 20221122 with $n_s=7$, where temporal decomposition is lower than spatial by 0.16 percentage points. Full results are reported in~\ref{app:ablation-direction}.

Across all six configurations, spatial decomposition achieves the lowest error in five of six comparisons against temporal decomposition and in four of six against space-time decomposition, for a total of 9 of 12 pairwise comparisons. The three pairwise comparisons in which spatial decomposition does not rank first all involve the normal congestion day (20221122) at higher sensor counts ($n_s = 5$ and $7$), where the differences among directions are below 0.20 percentage points. This record is an empirical result under the sensing configurations tested, and it does not establish an interaction with transition structure.

Mean error and training time separate the three directions differently. Spatial averages 17.02\% at 474~s, temporal 17.83\% at 339~s, and decomposition along both space and time 17.07\% at 998~s. Therefore, spatial decomposition is neither the fastest nor uniformly the most accurate. The space--time variant has a similar mean error but requires 2.1 times the training time, since it creates four subdomains rather than two. This pattern of accuracy and cost is consistent with allocating local capacity along the sparsely observed spatial direction, but the analytical motivation in Section~\ref{sec:direction} does not establish that spatial decomposition is optimal, either in these runs or in general.

\subsection{Sensor Sensitivity Analysis}\label{sec:results-sensor}

\begin{figure}[!htbp]
\centering
\includegraphics[width=0.6\columnwidth]{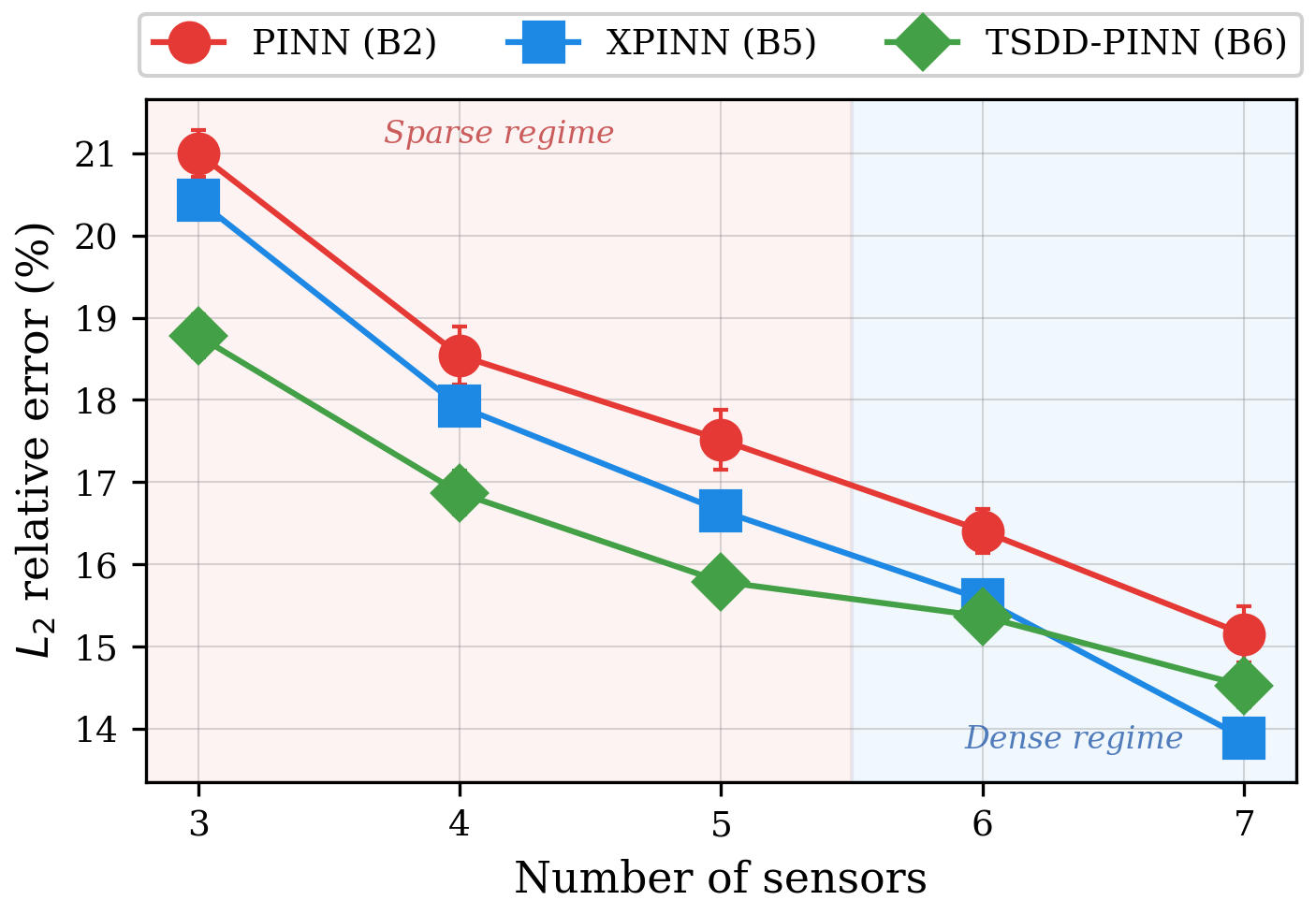}
\caption{Mean relative $L_2$ error versus sensor count for Vanilla PINN (B2), XPINN (B5), and TSDD-PINN (B6), averaged across the five I-24 days. Error bars use the original convention: population standard deviation across ten seeds within each day--method--sensor cell, averaged across the five days.}
\label{fig:sensor-sensitivity}
\end{figure}

Table~\ref{tab:sensor-sensitivity} and Figure~\ref{fig:sensor-sensitivity} summarize how performance changes with sensor density, averaged across all five I-24 days. {TSDD-PINN} consistently improves on Vanilla PINN, with the largest gain at the sparsest setting, $n_s=3$, where the mean reduction is 2.22 percentage points. The corrected gain sequence is 2.22, 1.83, 1.73, 1.04, and 0.63 percentage points for $n_s=3$ through 7.

\begin{table}[!htbp]
\centering
\caption{Mean relative $L_2$ error (\%) as a function of sensor count, averaged across all five I-24 datasets. $\Delta$ denotes the improvement of B6 over B2 in percentage points.}
\label{tab:sensor-sensitivity}
\small
\begin{tabular}{c c c c c}
\toprule
$n_s$ & B2 (PINN) & B5 (XPINN) & B6 (Ours) & $\Delta$ (B2$\to$B6) \\
\midrule
3 & 21.00 & 20.43 & \textbf{18.78} & $+2.22$ \\
4 & 18.70 & 17.92 & \textbf{16.87} & $+1.83$ \\
5 & 17.52 & 16.65 & \textbf{15.79} & $+1.73$ \\
6 & 16.41 & 15.57 & \textbf{15.37} & $+1.04$ \\
7 & 15.15 & \textbf{14.09} & 14.52 & $+0.63$ \\
\bottomrule
\end{tabular}
\end{table}

These results are relevant for freeway monitoring in under-instrumented corridors. In such settings, the estimator must reconstruct long unobserved spatial intervals from a small number of fixed-sensor traces. The widening gap in Figure~\ref{fig:sensor-sensitivity} indicates that {TSDD-PINN} provides the largest improvement in this evaluated sparse-sensing regime. The crossing at $n_s=7$ should be interpreted as a scope condition rather than a contradiction. Under denser virtual-sensor settings, the advantage of residual-guided spatial decomposition diminishes, and the fixed space-time partition of XPINN is no longer consistently disadvantaged.

Canonical time-slice spatial linear interpolation averages 16.51\%, and ASM averages 18.12\% after observation-only LOOCV selection of its spatial bandwidth. Linear interpolation is evaluated on all 7{,}200 time rows, whereas ASM is evaluated on its own 360-row grid; no evaluation-grid equivalence is claimed. The linear-interpolation comparison is not significant overall ($p=.443$): TSDD-PINN has lower error in 13 of 15 sparse cases at $n_s=3$--5, whereas interpolation has lower error in 8 of 10 dense cases. \ref{app:classical-details} gives the calibration details and paired statistics.

The primary normalization uses the complete-field minimum and maximum speeds, and the complete-field 95th-percentile $v_f$, in an offline virtual-sensor protocol. Across 75 matched sparse cases ($n=75$), sparse-only minus primary is $-0.13$ percentage points (95\% CI [$-0.35,+0.10$], $p=.262$), with a maximum $v_f$ shift of 0.59~mph. Nonsignificance does not establish equivalence or remove the complete-field information path. The one-at-a-time four-parameter sensitivity analysis covers $\tau_{\mathrm{shock}}=1.25$--$3.00$, peak fraction $0.15$--$0.60$, inter-peak distance $0.05$--$0.20$, and minimum width $0.075$--$0.30$. At $\tau_{\mathrm{shock}}=2.0$, the default-continuation mean is 16.69\%. Across the seven tested thresholds, the assembled means span 0.11 percentage points, although lower-threshold activations are partly represented by two-subdomain proxies rather than fully executed alternative policies. For 20221129 with $n_s=3$, the peak-fraction alternatives produce an adverse 24.67\%--41.24\% range. Minimum width does not guarantee the presence of sensor-populated children. Therefore, the analysis does not establish robustness across all local settings.

\subsection{Computational Efficiency}\label{sec:results-efficiency}

Table~\ref{tab:efficiency} compares mean accuracy and training time across all 25 I-24 configurations. {TSDD-PINN} achieves the best average relative $L_2$ error at 16.26\% while requiring 496 s of training time, compared with 1,189 s for XPINN. Under the evaluated implementations, subdomain structures, and training budgets, TSDD-PINN trains 2.4 times faster than XPINN (best baseline). Since network count, initialization, batching, and optimization path differ, the result is not attributed to warm start or any single component.

\begin{table}[!htbp]
\centering
\caption{Mean accuracy and training time on I-24 MOTION, averaged across all 25 configurations (5 days $\times$ 5 sensor counts) and 10 seeds (250 runs per method).}
\label{tab:efficiency}
\small
\begin{tabular}{l c c c}
\toprule
Method & $L_2$ (\%) & Time (s) & Subdomains \\
\midrule
B1 (NN) & 34.71 & 39 & 1 \\
B2 (PINN) & 17.75 & 538 & 1 \\
B3 (RAR) & 17.16 & 115 & 1 \\
B4 (Visc.) & 17.07 & 249 & 1 \\
B5 (XPINN) & 16.93 & 1{,}189 & 4 \\
B6 (Ours) & \textbf{16.26} & 496 & 2 \\
\bottomrule
\end{tabular}
\end{table}

PINN+RAR (B3) is the fastest physics-informed baseline because, in addition to focusing training on high-residual regions through adaptive resampling, it processes a per-step collocation mini-batch rather than the full collocation pool. However, this speed comes at the expense of lower accuracy than {TSDD-PINN}. Among the physics-informed methods, B3 has the shortest mean training time (115~s), whereas {TSDD-PINN} has the lowest mean error (16.26\%) and requires 496~s. Method choice depends on the relative importance of reconstruction error and training time.

\subsection{Negative-Control Experiment on NGSIM}\label{sec:results-ngsim}

The NGSIM experiment, described in Section~\ref{sec:datasets}, serves as a negative-control case on a second dataset for {TSDD-PINN}. It is not evidence that a transition regime is absent. The prespecified screen did not activate in any of the 50 historical NGSIM runs. The {TSDD-PINN} configuration (B6$^{\ast}$) retained the single-domain fallback. This no-trigger outcome does not validate selector quality or a transition-mechanism rule.

\begin{table}[!htbp]
\centering
\caption{Results of the NGSIM negative control. B6$^{\ast}$ denotes the existing continuation path after 50 no-trigger outcomes. B6$^{\dagger}$ denotes the forced-decomposition reference. Values are mean relative $L_2$ error (\%) over 10 seeds.}
\label{tab:ngsim}
\small
\begin{tabular}{c c c c}
\toprule
$n_s$ & B6$^{\ast}$ (Ours) & B5 (XPINN) & B6$^{\dagger}$ (Forced) \\
\midrule
3 & \textbf{18.75} & 19.14 & 19.67 \\
4 & \textbf{16.76} & 16.86 & 17.29 \\
5 & \textbf{15.35} & 15.51 & 16.26 \\
6 & \textbf{13.15} & 13.21 & 14.62 \\
7 & \textbf{12.25} & 12.45 & 13.94 \\
\bottomrule
\end{tabular}
\end{table}

Table~\ref{tab:ngsim} reports that B6$^{\ast}$ ranks first in all five sensor configurations, with an overall relative $L_2$ error of 15.25\% and a mean training time of 123 s, compared with 15.44\% and 668 s for XPINN. By contrast, B6$^{\dagger}$ degrades to 16.36\% and 273 s, ranking sixth of seven methods overall.

The contrast between B6$^{\ast}$ and B6$^{\dagger}$ shows a lower-error existing continuation beside an adverse forced-decomposition reference; it does not identify why the screen did not activate or validate its decision. This is a negative control, not validation of the activation threshold on I-24. Therefore, NGSIM narrows the cross-dataset claim.

\section{Discussion}\label{sec:discussion}

The results support a narrower claim than \emph{domain decomposition always helps PINNs}. The controlled evidence for the combined refinement method is concentrated where fixed sensors leave long spatial gaps; dense interpolation and NGSIM results delimit that operating scope. This interpretation is consistent with prior work on TSE observability, DD-PINN generalization, and PINN failure on transport-dominated PDEs \citep{seo2017traffic,hu2021extended,huang2023limitations,karniadakis2021physics}.

\subsection{Why Decomposition Direction Matters for Fixed-Sensor TSE}\label{sec:disc-spatial}

Three observations bound the interpretation of the findings of Section~\ref{sec:results-ablation}. First, spatial splitting can reduce the local approximation burden created by spectral bias, gradient imbalance, and shock-prone LWR structure, consistent with PINN and conservation-law studies \citep{tancik2020fourier,wang2021understanding,krishnapriyan2021characterizing,huang2023limitations,lei2025potential,coutinho2023physics,deryck2024wpinns,lorin2024nondiffusive,neelan2025physics,wang2024respecting,wu2023comprehensive}. Second, fixed sensors provide dense temporal traces but sparse spatial coverage, so a spatial split aligns local capacity with the unobserved intervals \citep{seo2017traffic,wang2005real,wang2008real,zhao2024observer,hu2024sensor}. Third, the matched direction comparison shows that spatial refinement attains lower mean error than temporal refinement under the tested fixed-sensor conditions. The coarse residual is used only to guide a deterministic partition and is not treated as evidence of spatial error dominance. This scope matches DD-PINN analyses showing that decomposition is effective when it simplifies local functions sufficiently to outweigh interface costs and the reduced data per subdomain \citep{hu2021extended,shukla2021parallel,moseley2023finite}. Neither the observation result nor the residual profile proves the empirical direction result.

The I-24 and NGSIM results define a sensing-density- and dataset-dependent scope, not a transition-mechanism rule. The largest gains occur on the severe and incident I-24 days, whereas the NGSIM shock-screened mode retained the single-domain fallback in all 50 runs and ranked first across sensor configurations. When forced decomposition was applied to NGSIM, the same architecture dropped to sixth of seven methods overall. This adverse forced result narrows the cross-regime claim but does not validate the screen as a selector \citep{huang2022physics,shi2021a,shi2021b,lu2023physics}.

XPINN is a useful DD baseline because it supports spatial, temporal, and space-time partitions within one unified framework \citep{jagtap2020extended}. The observed gains are consistent with a combined design comprising residual-guided split placement, a minimal two-subdomain structure, parent-to-child warm start, and fixed-interface regularization. Therefore, the contribution of {TSDD-PINN} should be interpreted as a traffic-specific decomposition workflow rather than as a claim that each component is new in isolation. The completed controls find borderline initialization evidence concentrated in one day, a null result for exact valley placement, and no activation of the conditional R-H/entropy branch.

\subsection{Traffic Engineering Implications}\label{sec:disc-implications}

{TSDD-PINN} is complementary to classical TSE methods. Canonical linear interpolation is often lower under dense sensing, and observation-matched ASM uses published fixed parameters and observed-trace LOOCV bandwidth selection. Extended Kalman filtering has supported real-time freeway traffic-state estimation and online parameter estimation under detector layouts that include boundary and ramp information. Hamilton--Jacobi formulations provide a physically interpretable framework for assimilating boundary, internal, cumulative-count, and mobile-sensor information. The present benchmark instead supplies equally spaced interior virtual speed traces without measured boundary or ramp flows or cumulative counts; implementing these methods would change the observation and model specification rather than provide a matched comparison \citep{wang2005real,wang2008real,treiber2002reconstructing,daganzo2006variational,canepa2017networked}, while {TSDD-PINN} targets the evaluated offline reconstruction setting with sparse fixed sensors.

From a practical implementation perspective, classical model-based estimators provide interpretable traffic states and can support recursive real-time operation once the traffic model, boundary conditions, measurement equations, and uncertainty specifications are established. Their application may nevertheless require corridor-specific calibration, boundary or ramp measurements, and expert specification of process and measurement models. TSDD-PINN instead fits sparse speed observations and the LWR residual within one differentiable model, shifting part of the fitting process into training without eliminating expert choices: the current implementation still requires normalization statistics, a prescribed $v_f$, network and loss settings, screening and partition parameters, and offline optimization. Classical methods may be preferable for calibrated online operation, whereas TSDD-PINN is presently positioned for offline reconstruction from sparse interior speed traces. In the offline I-24 reconstruction setting, estimator design can shift the sensor-density/error trade-off. {TSDD-PINN} with 4 sensors attains 16.87\% relative $L_2$ error, lower than Vanilla PINN with 5 sensors at 17.52\%, and {TSDD-PINN} with 5 sensors attains 15.79\%, lower than Vanilla PINN with 6 sensors at 16.41\%. In these equally spaced virtual-sensor experiments, comparable errors are reached with one fewer interior sensor in the examples above. This should not be interpreted as a direct recommendation for sensor deployment; rather, it suggests that model design can affect the cost-accuracy frontier in offline reconstruction.

Current training times are better suited to post hoc corridor reconstruction, incident forensics, daily monitoring, and archived reconstruction than to streaming estimation \citep{seo2017traffic,treiber2002reconstructing,lu2023physics,rempe2022estimation}. The optional no-trigger rule prevents further training on no-trigger paths, but its derived estimate was obtained without an additional full training experiment and is not a universal improvement guarantee.

\subsection{Limitations and Future Work}\label{sec:disc-limitations}

The first limitation is model scope. We use LWR with a Greenshields closure and a fixed, pre-estimated $v_f$ to isolate the domain-decomposition design in a simple setting where shock-related PINN failure is documented \citep{huang2023limitations}. This choice does not imply that the Greenshields closure or first-order deterministic physics is sufficient for all TSE tasks. Second-order, stochastic, heterogeneous, and jointly learned constitutive models may be more realistic for settings with lane-changing, ramps, or class-specific behavior and remain natural extensions beyond the present evaluation \citep{aw2000resurrection,aw2002derivation,zhang2002,shi2021a,shi2021b,wong2002,jabari2012,jabari2013stochastic,yuan2021,wang2022real}.

A second limitation is numerical: the valley-based split rule is heuristic, and a low-residual valley may not provide the optimal interface placement when shocks move substantially over time. Localization is not significantly better than the tested midpoint and random controls, and physical separation fails in four of six evaluated cases. Because the interface formulation is fixed-interface regularization, not a localization or exact shock-tracking method, future work should combine domain-decomposition refinement with weak-form or discontinuity-aware losses \citep{deryck2024wpinns,lorin2024nondiffusive,neelan2025physics}.

A complementary future direction is to formulate the LWR prior in the cumulative-count variable $N(x,t)$ rather than directly in $(q,\rho)$ or speed. Newell's cumulative-flow construction and Daganzo's variational formulation show that kinematic-wave solutions can be represented through a continuous, piecewise differentiable cumulative-count function satisfying a Hamilton-Jacobi structure, so shocks appear as slope discontinuities rather than state-variable jumps \citep{newell1993general,daganzo2005variational}. Such a formulation may reduce the direct conflict between continuous neural approximators and discontinuous traffic states, but it would also change the observation model because fixed sensors provide point measurements of speed, flow, or occupancy rather than cumulative counts. Residual-guided split detection and parent-to-child warm start would remain conceptually complementary in that setting because they operate on learned residual fields and network weights rather than on a specific choice of traffic state variable.

A third limitation is data scope and sensing mode. The formulation is lane-aggregated, single-corridor, and source-free, so ramp-influenced cells may not be fully represented by the LWR residual. Therefore, the evidence should be interpreted as same-corridor multi-day robustness, not cross-roadway generalization. The use of full-field normalization statistics and a pre-estimated $v_f$ also fixes the present study as an offline reconstruction evaluation rather than an online deployment study. The sparse-only normalization difference is nonsignificant, but this does not establish equivalence or remove the offline information path; the complete-field information path may overestimate practical deployment performance. Broader validation is needed across corridors, incident types, ramps, junctions, and sensing mixes. Probe vehicles and other Lagrangian data are especially important because richer spatial coverage may alter observability and the preferred decomposition direction \citep{herrera2010incorporation,seo2015probe,canepa2017networked,rempe2022estimation,wang2022real}. Network-scale extensions require coordinated link partitions and ramp source terms \citep{usama2022physics,lu2023physics}.

A fourth limitation is benchmark scope. Canonical interpolation is observation- and evaluation-grid matched; ASM is observation-matched on its own 360-row grid. Under the present equally spaced interior-speed virtual-sensor protocol, a matched extended Kalman filtering or variational Hamilton--Jacobi implementation would require additional observations or model specifications, including measured boundary or ramp flows when available, cumulative-count constraints, and calibrated state-transition and measurement models. Published studies document successful applications of these methods under observation settings that include such information, but their reported numerical results are not directly commensurate with the present full-field relative-$L_2$ evaluation \citep{wang2005real,wang2008real,treiber2002reconstructing,daganzo2006variational,canepa2017networked}.  The architecture, loss weights, shock threshold, and split-detection thresholds are fixed protocol choices for fair relative comparison, not optimized deployment prescriptions. The sensitivity analysis does not establish robustness in every local setting. Future work should also compare with operator-learning approaches and incorporate uncertainty quantification using Bayesian or uncertainty-aware PINN tools \citep{zhang2019quantifying,yang2021bpinns}.

\section{Conclusion}\label{sec:conclusion}

This study proposed {TSDD-PINN}, an observation-aligned two-stage domain-decomposition framework for offline freeway reconstruction from sparse fixed sensors. The framework supports spatial, temporal, and space--time variants. Among the evaluated directions, spatial refinement attains the lowest mean error and requires less than half the training time of space--time refinement, while temporal refinement is faster. The method first trains a coarse global PINN, separates a controlled refinement path from an optional Stage-1-preserving operational path, uses the residual profile as a deterministic partition heuristic when spatial decomposition is enabled, initializes child networks from the parent solution, and applies fixed-interface regularization.

The results support four main findings. First, controlled {TSDD-PINN} achieved the lowest relative $L_2$ error in 18 of 25 I-24 configurations and in 14 of 15 sparse-sensing cases, while the overall comparison with linear interpolation is not significant, and dense sensing is often adequately served by interpolation. The completed default-screen evaluation records 20/250 activations and 16.69\% for the existing continuation; the derived no-trigger estimate of 16.3873\% was obtained without an additional full training experiment and does not validate the screen. Second, spatial decomposition was lower in 9 of 12 direction comparisons for fixed-sensor TSE in the studied regimes. The space--time variant has a similar mean error but requires 2.1 times the training time, while temporal decomposition was faster. Third, under the evaluated implementations, subdomain structures, and training budgets, TSDD-PINN had lower mean error than XPINN (16.26\% versus 16.93\%) and required 496~s versus 1{,}189~s on I-24. Fourth, the evaluation covered five I-24 days, five sensor configurations, and ten seeds, yielding 1{,}500 I-24 runs across six methods.

Overall, {TSDD-PINN} has a supported operating range that depends on sensing density rather than an advantage that holds across settings. The method should be interpreted as a PINN-family framework for offline freeway reconstruction rather than as a replacement for calibrated classical TSE estimators or as an online deployment method. Residual localization and exact valley placement have not been validated as separate contributors to performance. The conditional Rankine--Hugoniot/entropy branch did not activate during the component analysis, and the NGSIM result is inconsistent with a benefit from the no-trigger path that holds across settings. Future work should extend {TSDD-PINN} to richer traffic physics, network-scale corridors with ramps and junctions, online normalization and parameter estimation, matched comparisons with classical TSE estimators, uncertainty quantification, cumulative-count formulations, and weak-form treatments of shocks.

\section*{CRediT authorship contribution statement}
\textbf{Eunhan Ka}: Conceptualization, Methodology, Software, Investigation, Writing -- original draft, Visualization. \textbf{Ludovic Leclercq}: Investigation, Writing -- review \& editing. \textbf{Satish V. Ukkusuri}: Conceptualization, Methodology, Investigation, Resources, Writing -- review \& editing, Supervision.

\section*{Acknowledgments}
This work is based upon the work supported by the National Center for Transportation Cybersecurity and Resiliency (TraCR) (U.S. Department of Transportation National University Transportation Center) headquartered at Clemson University, Clemson, South Carolina, USA. Any opinions, findings, conclusions, and recommendations expressed in this material are those of the author(s) and do not necessarily reflect the views of TraCR, and the U.S. Government assumes no liability for the contents or use thereof.

\section*{Declaration of competing interest}
The authors declare that they have no known competing financial interests or personal relationships that could have appeared to influence the work reported in this paper.

\section*{Data availability}
The full source code, processed datasets, and experimental configurations used in this study will be released at \url{https://github.com/eunhanka/add-pinn} upon acceptance. The I-24 MOTION trajectory data are available from \citet{gloudemans202324}. The NGSIM I-80 trajectory data are publicly available from the U.S.\ Department of Transportation Federal Highway Administration \citep{usdot_ngsim_2016}; the processed NGSIM protocol follows prior traffic PINN studies \citep{huang2022physics}.

\section*{Declaration of generative AI use}
During the preparation of this work, the authors used ChatGPT (OpenAI) in order to assist with language refinement and LaTeX formatting. After using this tool, the authors reviewed and edited the content as needed and take full responsibility for the published article.

\bibliographystyle{elsarticle-harv}
\bibliography{reference}

\newpage
\appendix

\FloatBarrier

\section{Notation, Analytical Assumptions, and Proofs}\label{app:theory-proofs}
\subsection{Notation}\label{app:notation}

\begin{table}[!htbp]
\footnotesize
\centering
\caption{Notation used in the manuscript.}
\label{tab:notation}
\begin{tabularx}{\textwidth}{@{}lX@{\quad}lX@{}}
\toprule
Symbol & Description & Symbol & Description \\
\midrule
\multicolumn{4}{@{}l}{\textit{Traffic flow variables and domain}} \\
$u(x,t)$ & Speed field (mph) & $\Omega$ & Spatiotemporal domain \\
$\rho(x,t)$ & Traffic density (vehicles/mile) & $\hat{x},\hat{t},\hat{u}$ & Normalized coordinates and speed \\
$\hat{\rho}$ & Normalized density analog & $x_{\min},x_{\max}$ & Spatial domain bounds (feet) \\
$v_f$ & Free-flow speed (mph) & $u_{\min}$ & Physical minimum speed (mph) \\
$\rho_{\mathrm{jam}}$ & Physical jam density (vehicles/mile) & $u_{\max}$ & Physical maximum speed (mph) \\
$\hat{\rho}_{\mathrm{jam}}$ & Normalized jam density ($=1$) & $\hat{u}_{\max}$ & Normalized maximum speed ($=1$) \\
$T$ & Temporal domain extent (seconds) & $N$ & Number of spatial cells \\
$q(\rho)$ & Physical flow & $\hat{q}(\hat{\rho})$ & Normalized interface flux \\
$n_s$ & Number of fixed sensors & & \\
\midrule
\multicolumn{4}{l}{\textit{Observation and idealized representation}}\\

\(S=\{x_i\}_{i=1}^{n_s}\)
& Fixed-sensor locations
& \(\alpha\)
& Relative jump position within a sensor gap\\

\(C^1(\Omega)\)
& Continuously differentiable fields on \(\Omega\)
& \(h=x_{i+1}-x_i\)
& Gap between adjacent sensors \\

\(u_L,u_R,\Delta u\)
& Left/right plateau speeds and jump \(\Delta u=u_R-u_L\)
& \(\mathcal{H}_S\)
& Fixed-sensor temporal-trace operator;
  \(\mathcal{H}_S u\) is the trace vector for field \(u\) \\

\(I_h u\)
& Endpoint linear interpolant over a gap \(h\)
& \(\xi,\widehat{\xi}\)
& True and represented jump locations \\

\(\widehat{u}_{\widehat{\xi}}\)
& Idealized two-piece representation
& \(\lVert\cdot\rVert_{L^2(x_i,x_{i+1})}\)
& Spatial \(L^2\) norm over a sensor gap \\
\midrule
\multicolumn{4}{@{}l}{\textit{PDE, residual, and decomposition}} \\
$r(x,t)$ & Normalized PDE residual & $n$ & Number of subdomains \\
$A,B,C$ & Nondimensionalization coefficients & $\theta$ & Network parameters \\
$R(x), R(t)$ & Spatial / temporal residual profiles & $\mathbf{W}$ & Fourier feature matrix \\
$n_x, n_t$ & Residual evaluation grid dimensions & $d_e$ & Fourier embedding dimension \\
$K$ & Smoothing kernel size & $\sigma$ & Fourier feature scale \\
\midrule
\multicolumn{4}{@{}l}{\textit{Loss functions and training}} \\
$\mathcal{L}_{\mathrm{total}}$ & Total training loss & $w_{\mathrm{data}}, w_{\mathrm{pde}}, w_{\mathrm{int}}$ & Fixed loss weights \\
$\mathcal{L}_{\mathrm{data}}$ & Data loss on sensor observations & $w_j^{\mathrm{causal}}$ & Causal weighting factor \\
$\mathcal{L}_{\mathrm{pde}}$ & PDE residual loss & $\epsilon$ & Causality parameter \\
$\mathcal{L}_{\mathrm{int}}$ & Interface coupling loss & $E_{\mathrm{split}}$ & Split epoch for decomposition \\
$\mathcal{O}$ & Sensor observation set & $E_{\max}$ & Maximum training epochs \\
$N_{\mathrm{data}}$ & Number of sensor observations & $N_{\mathrm{pde}}^s$ & Collocation points in subdomain $s$ \\
\midrule
\multicolumn{4}{@{}l}{\textit{Interface quantities}} \\
$s$ & Learned dimensionless interface speed & $\delta_{\mathrm{shock}}$ & Conditional-interface threshold \\
$\mathcal{L}_{\mathrm{RH}}$ & Rankine--Hugoniot interface loss & $\lambda(\rho),\hat{\lambda}(\hat{\rho})$ & Physical and normalized characteristic speeds \\
$\mathcal{L}_{\mathrm{entropy}}$ & Entropy admissibility loss & & \\
\bottomrule
\end{tabularx}
\end{table}

\subsection{Assumptions and proofs}\label{app:theory-details}

The observation statement assumes a nondegenerate rectangle \(\Omega=[x_{\min},x_{\max}]\times[0,T]\), with \(x_{\min}<x_{\max}\) and \(T>0\), and a finite set \(S=\{x_1,\ldots,x_{n_s}\}\) of distinct fixed locations, indexed so that \(x_1<\cdots<x_{n_s}\). Each detector record is idealized as a complete temporal trace, and the argument is posed on \(C^1(\Omega)\), the space of continuously differentiable speed fields on \(\Omega\). This smooth space is used only for the observation-operator statement. The latter representation identities instead use a separate fixed-time, piecewise-constant slice; that discontinuous slice is not assumed to belong to \(C^1(\Omega)\).

The fixed-sensor temporal-trace operator introduced in
Section~\ref{sec:problem-formulation} is
\begin{equation}\label{eq:app-observation}
\mathcal H_S:
C^1(\Omega)\to[C^1([0,T])]^{n_s},
\qquad
\mathcal H_S u
=
\bigl(u(x_1,\cdot),\ldots,u(x_{n_s},\cdot)\bigr),
\end{equation}
where \([C^1([0,T])]^{n_s}\) denotes ordered \(n_s\)-tuples of
continuously differentiable temporal traces.

\paragraph{Fixed-sensor observation ambiguity}
Define
\[
p(x)=\prod_{j=1}^{n_s}(x-x_j),
\]
choose a fixed nonzero $\psi\in C^1([0,T])$, and set
\[
v_k(x,t)=p(x)(x-x_{\min})^k\psi(t),\qquad k=0,1,\ldots.
\]
Every $v_k$ belongs to $C^1(\Omega)$. Because $p(x_j)=0$ for each
detector, $v_k(x_j,t)=0$ for all $t$, and hence
$\mathcal H_S v_k=0$. To establish linear independence, suppose that $\sum_{k=0}^{m}c_kv_k=0$. Select a time $t_0$ for which $\psi(t_0)\ne0$ and an open spatial interval containing no sensor. On that interval $p(x)\ne0$, so division by $p(x)\psi(t_0)$ gives
\[
\sum_{k=0}^{m}c_k(x-x_{\min})^k=0.
\]
A polynomial that vanishes on an open interval is identically zero,
so every \(c_k\) is zero. Thus, \(\{v_k\}_{k\ge0}\) is an infinite
linearly independent family satisfying
\(\mathcal H_S v_k=0\). Hence, infinitely many independent field
variations are invisible to the fixed-sensor temporal traces. This
conclusion concerns \(\mathcal H_S\) alone: it neither establishes
nonuniqueness of the fully LWR-constrained reconstruction nor shows
that domain decomposition removes this observation ambiguity.

\paragraph{One-jump endpoint interpolation}
Fix a time and let adjacent sensors at $x_i$ and $x_{i+1}$ be separated by $h=x_{i+1}-x_i>0$. Assume a noiseless piecewise-constant slice with exact endpoint plateaus $u_L$ and $u_R$ and one jump $\Delta u=u_R-u_L$ at $\xi=x_i+\alpha h$, where $0<\alpha<1$. With $y=x-x_i$, write
\[
u(x_i+y)=
\begin{cases}
u_L, & 0\le y<\alpha h,\\
u_R, & \alpha h\le y\le h,
\end{cases}
\qquad
I_h u(x_i+y)=u_L+\Delta u\,\frac{y}{h}.
\]
On the left of the jump, the error magnitude is $|\Delta u|y/h$; on the right it is $|\Delta u|(1-y/h)$. Direct integration gives
\begin{align}
\lVert u-I_h u\rVert_{L^2}^2
&=(\Delta u)^2\left[\int_0^{\alpha h}\frac{y^2}{h^2}\,dy+
\int_{\alpha h}^{h}\left(1-\frac yh\right)^2\,dy\right] \\
&=\frac{(\Delta u)^2h}{3}\{\alpha^3+(1-\alpha)^3\}
\ge\frac{(\Delta u)^2h}{12}.
\label{eq:app-interpolation}
\end{align}
The inequality follows from
\[
\alpha^3+(1-\alpha)^3=3\left(\alpha-\tfrac12\right)^2+\tfrac14\ge\tfrac14.
\]
It describes the spatial $L^2$ representation error of endpoint linear interpolation for this idealized one-jump slice. The scaling with $h$ formalizes a gap burden, but it is not a prediction for a trained neural estimator.

\paragraph{Fixed split representation}
Under the same fixed-time assumptions, let $\widehat\xi\in[x_i,x_{i+1}]$ and define an idealized two-piece representation $\widehat u_{\widehat\xi}$ using the exact plateau states $u_L$ and $u_R$ but placing the interface at $\widehat\xi$. The target and representation agree outside the interval between $\xi$ and $\widehat\xi$ and differ there by $|\Delta u|$. Therefore,
\begin{equation}\label{eq:app-split}
\lVert u-\widehat u_{\widehat\xi}\rVert_{L^2}^2
=\int_{\min(\xi,\widehat\xi)}^{\max(\xi,\widehat\xi)}(\Delta u)^2\,dx
=(\Delta u)^2|\widehat\xi-\xi|.
\end{equation}

Both identities assume one fixed-time jump, exact plateau values, no observation noise, and a spatial $L^2$ norm. They invoke neither the LWR constraint nor a neural-network hypothesis class. They are not lower bounds on TSDD-PINN error, convergence results, guarantees of learned-interface accuracy, or evidence that a deployed residual valley is optimal. In particular, the split identity uses the true plateau states and perturbs only the interface; the implemented method must estimate a full field and select its interface from a coarse-model residual. The identities do not prove empirical or universal spatial superiority, justify any direction as universally optimal, or validate the actual residual-selected split. They provide only a bounded analytical motivation for testing spatial refinement under the fixed-sensor geometry; the empirical comparisons, not these idealizations, determine the reported operating scope.

\clearpage
\section{Data-Site Context and Training Procedure}\label{app:default-evaluation}

\subsection{Data-site context}\label{app:data-sites}

\begin{figure}[!htbp]
    \centering
    \includegraphics[width=0.9\textwidth]{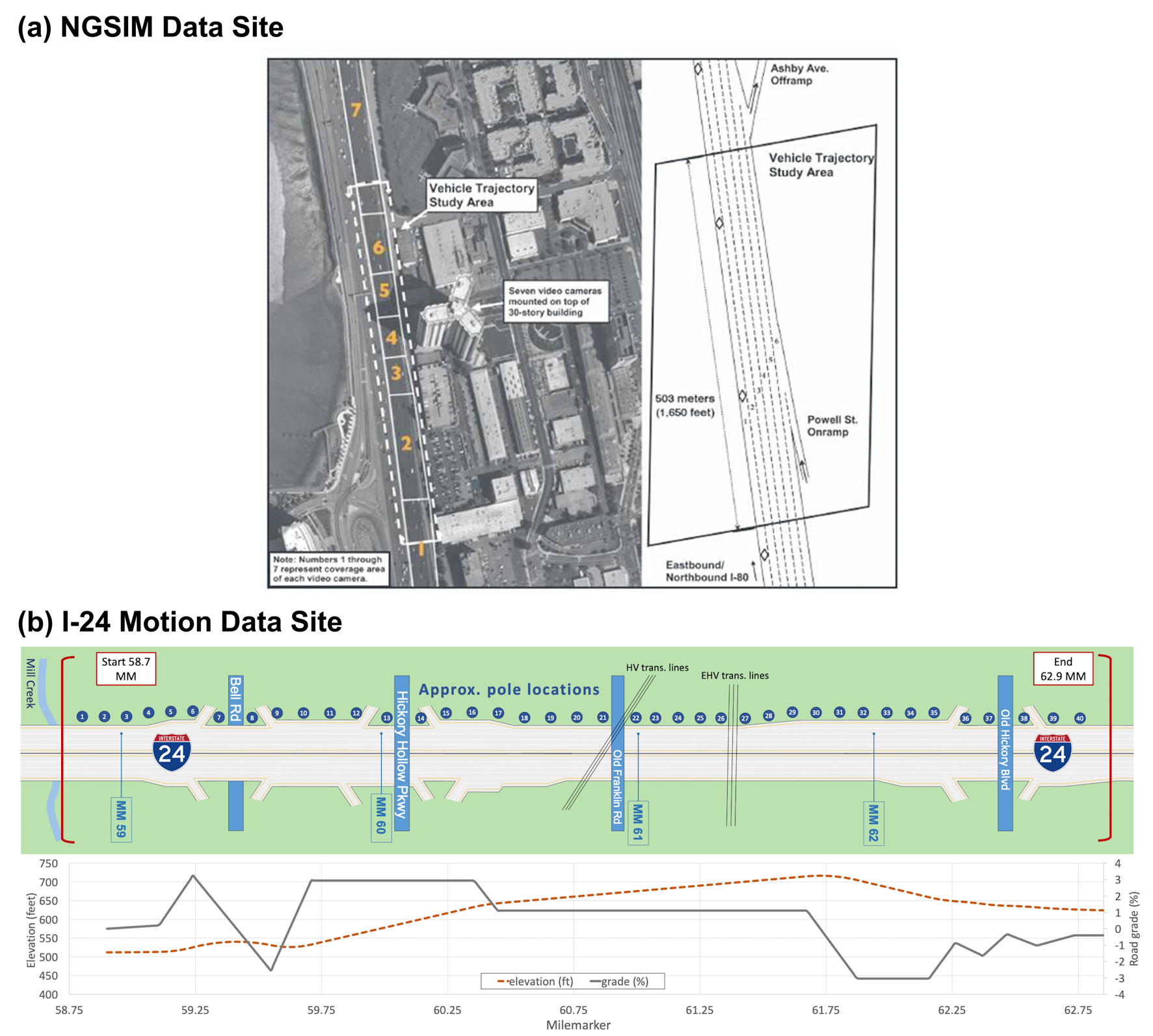}
    \caption{Vehicle trajectory data collection sites used in this study.
    \textbf{(a)} NGSIM I-80 trajectory study area in Emeryville, California, covering a 1{,}600-ft freeway segment.
    \textbf{(b)} I-24 MOTION collection site along Interstate~24 near Nashville, Tennessee, covering the 4.2-mile corridor used in the primary experiments.
    (Sources: \citet{usdot_ngsim_2016} and \citet{gloudemans202324}).}
    \label{fig:trajectory-datasets}
\end{figure}

\subsection{Training and evaluation paths}\label{app:evaluation-paths}

The controlled path prespecifies refinement and uses the residual-selected spatial partition; it supplies the I-24 benchmark in Table~\ref{tab:main-results} and is not the output of the default screen. The separate prespecified default-screen evaluation completed 250/250 I-24 runs. It activated in 20/250 runs from 2/25 configurations; the aggregate means were 16.69\% for the existing default continuation, 16.3893\% for captured Stage~1, and 16.2648\% for controlled refinement. Because these paths also differ in architecture, distillation, RAR, and interface operations, the comparison does not validate the screen.

The Stage-1-preserving no-trigger rule returns the captured Stage~1 predictor before any further training when the screen does not activate; in triggered cases, it follows the same refinement procedure as the controlled path. Its derived I-24 mean is 16.3873\%. No additional full training experiment was conducted for this estimate, and it does not establish selector validity or universal non-inferiority.

\subsection{TSDD-PINN training procedure}\label{app:algorithm}

Algorithm~\ref{alg:training} separates the controlled path from the optional operational safeguard. The no-trigger return occurs before any additional refinement or training, and the activated case follows the same refinement procedure as the controlled path.
\vspace{-1pt}
\begin{algorithm}[H]

\small
\caption{TSDD-PINN Two-Stage Training}
\label{alg:training}
\begin{algorithmic}[1]

\REQUIRE Sensor data $\mathcal{O}$; domain $\Omega$; $E_{\mathrm{split}}$; $E_{\max}$; $\tau_{\mathrm{shock}}$; controlled or operational evaluation
\ENSURE Captured Stage~1 predictor or refined spatial-subdomain predictor
\STATE Initialize the parent, LHS points, and fixed loss weights; train through $E_{\mathrm{split}}$ with causal data/PDE loss and Adam
\STATE Capture Stage~1 predictor $\widehat u_{\mathrm{S1}}$ and its state; compute $\mathcal{S}(\mathcal{O})$ from Equation~\eqref{eq:shock-indicator}
\IF{the evaluation is controlled or $\mathcal{S}(\mathcal{O})>\tau_{\mathrm{shock}}$}
    \STATE Compute $R(x)$; select width-constrained interfaces; initialize child networks from the parent
    \STATE Train through $E_{\max}$ with subdomain/interface losses, clipping, and RAR; return the refined predictor
\ELSE
    \RETURN $\widehat u_{\mathrm{S1}}$ \COMMENT{Stage-1-preserving no-trigger rule}
\ENDIF
\end{algorithmic}
\end{algorithm}

\section{Statistical and Classical-Comparator Details}\label{app:classical-details}

\subsection{Statistical robustness for I-24 pairwise comparisons}\label{app:statistical}

\begin{table}[H]
\centering
\caption{Pairwise statistics for TSDD-PINN (B6) versus five baselines across 25 I-24 configuration means. W/L: B6 wins/losses; $\Delta$: mean relative-$L_2$ improvement (percentage points) with 95\% CI. The paired $t$ and Wilcoxon tests are one-sided for lower B6 error; $p_{\mathrm{Holm}}$ adjusts the five paired $t$-tests by Holm--Bonferroni; $d$: paired Cohen's $d$.}
\label{tab:statistical-appendix}
\small
\setlength{\tabcolsep}{3.5pt}
\begin{tabularx}{\textwidth}{@{}l c X c c c c@{}}
\toprule
Comparison & W/L & \centering $\Delta$ [95\% CI] (pp) & $p_t$ & $p_W$ & $p_{\mathrm{Holm}}$ & $d$ \tabularnewline
\midrule
B6 vs B1 (NN) & 25/0 & \centering $+18.44$ [$16.55, 20.34$] & ${<}.0001$ & ${<}.0001$ & ${<}.0001$ & 4.02 \tabularnewline
B6 vs B2 (PINN) & 25/0 & \centering $+1.49$ [$1.15, 1.83$] & ${<}.0001$ & ${<}.0001$ & ${<}.0001$ & 1.80 \tabularnewline
B6 vs B3 (RAR) & 19/6 & \centering $+0.90$ [$0.52, 1.27$] & ${<}.0001$ & ${<}.0001$ & ${<}.0001$ & 0.99 \tabularnewline
B6 vs B4 (Visc.) & 19/6 & \centering $+0.80$ [$0.44, 1.17$] & ${<}.0001$ & $.0001$ & $.0001$ & 0.90 \tabularnewline
B6 vs B5 (XPINN) & 18/7 & \centering $+0.67$ [$0.29, 1.04$] & $.0006$ & $.001$ & $.0006$ & 0.73 \tabularnewline
\bottomrule
\end{tabularx}
\end{table}

\begingroup
\widowpenalty=0
\clubpenalty=0

\paragraph{Classical-comparator protocols and calibration}\label{app:classical-calibration}

For selected sensor $s$ at location $x_s$, let $v_s(t_i)$ denote its speed trace at time row $t_i$. The canonical time-slice spatial linear-interpolation comparator reconstructs each time row independently. For adjacent selected sensors $x_j<x_{j+1}$,
\[
u_{\mathrm{lin}}(x,t_i)
=
\frac{x_{j+1}-x}{x_{j+1}-x_j}\,v_j(t_i)
+
\frac{x-x_j}{x_{j+1}-x_j}\,v_{j+1}(t_i),
\qquad
x_j\leq x\leq x_{j+1}.
\]
Outside $[x_1,x_{n_s}]$, it holds the nearest selected-sensor value. The comparator uses all 7{,}200 I-24 time rows and is evaluated on the same 100-cell spatial grid as the neural estimators.

ASM reconstructs separate free-flow and congested fields using anisotropic space--time kernels and combines them through a speed-dependent transition weight. Let $r\in\{\mathrm{free},\mathrm{cong}\}$ denote the traffic regime, let $\sigma_x$ denote the spatial bandwidth, and let $\widehat v_{j,-j}^{(\sigma_x)}$ denote the reconstruction at held-out sensor $j$ using all other selected sensors. The reconstruction and observation-only bandwidth selectors are
\endgroup
\begin{equation}\label{eq:asm-appendix}
\begin{aligned}
w_s(x)&=e^{-|x-x_s|/\sigma_x},\quad
t_{q,s}^{r}=t-(x-x_s)/c_r,\quad
K_{si}^{r}=e^{-|t_{q,s}^{r}-t_i|/\tau_{\mathrm{ASM}}},\\
V_r(x,t)&=
\frac{\sum_s w_s\sum_i K_{si}^{r}v_s(t_i)}
{\sum_s w_s\sum_i K_{si}^{r}},\\
w_{\mathrm{cong}}&=
\tfrac12\!\left[
1+\tanh\!\left(
\frac{V_c-\min(V_{\mathrm{free}},V_{\mathrm{cong}})}
{\Delta V}
\right)
\right],\\
V_{\mathrm{ASM}}&=
w_{\mathrm{cong}}V_{\mathrm{cong}}
+
(1-w_{\mathrm{cong}})V_{\mathrm{free}},\\
E_{\mathrm{LOOCV}}(\sigma_x)&=
\frac{
\sqrt{\sum_j\sum_i
[\widehat v_{j,-j}^{(\sigma_x)}(t_i)-v_j(t_i)]^2}
}{
\sqrt{\sum_j\sum_i v_j(t_i)^2}
}.
\end{aligned}
\end{equation}
The published Treiber--Helbing values
$c_{\mathrm{free}}=49.7097$ mph,
$c_{\mathrm{cong}}=-9.3206$ mph,
$\tau_{\mathrm{ASM}}=0.0200$ h,
$V_c=31.0686$ mph, and
$\Delta V=6.2137$ mph are fixed
\citep{treiber2002reconstructing}. The two values considered for $\sigma_x$ are
0.3728 mi and one half of the selected-sensor mean spacing. The value
with the lower pooled LOOCV error across the held-out selected-sensor traces
is retained. ASM is then recomputed using all selected sensors before
the complete field is evaluated; no unobserved cell informs bandwidth selection.

\begin{table}[!htbp]

\centering
\caption{Classical comparator results: mean relative $L_2$ error (\%). Sparse is $n_s=3$--5 and dense is $n_s=6$--7. ASM denotes the adaptive smoothing method with observation-only LOOCV selection of the spatial bandwidth.}
\label{tab:classical-scope}
\small
\begin{tabular}{l c c c c}
\toprule
Method & Time rows & Overall & Sparse & Dense \\
\midrule
TSDD-PINN & 7{,}200 & \textbf{16.26} & \textbf{17.15} & 14.94 \\
Time-slice linear interpolation & 7{,}200 & 16.51 & 18.27 & \textbf{13.87} \\
ASM (observation-only LOOCV) & 360 & 18.12 & 19.16 & 16.56 \\
\bottomrule
\end{tabular}
\end{table}

For linear-minus-controlled $\Delta$, the overall comparison is 15W/10L with a mean difference of $+0.24$ percentage points (95\% CI [$-0.40,+0.89$], $p_t=.443$, $p_W=.491$). The sparse comparison is 13W/2L with $+1.12$ points (95\% CI [$+0.51,+1.74$], Holm $p_t=.005$, $p_W=.005$, $d=1.01$), whereas the dense comparison is 2W/8L with $-1.08$ points (95\% CI [$-1.91,-0.24$], Holm $p_t=.034$, $p_W=.074$, $d=-0.93$). The full-field ASM selector gives 17.95\% and is reported as an upper reference on comparator performance, not as a method that could be deployed. Linear interpolation uses 7{,}200 time rows, whereas ASM uses its own 360-row grid; no grid-equivalence claim is made. Under the present interior-speed virtual-sensor protocol, a matched extended Kalman filter or Hamilton--Jacobi estimator would require additional observations or model specifications, including measured boundary or ramp flows when available, cumulative-count constraints, and calibrated state-transition and measurement models.

\FloatBarrier
\section{Full Decomposition-Direction Results}\label{app:ablation-direction}

The direction comparison spans 180 results: 120 newly trained temporal and space--time runs and 60 spatial runs reused from controlled cells. Spatial, temporal, and space--time means are 17.02\%/474~s, 17.83\%/339~s, and 17.07\%/998~s; spatial is lower in 9/12 cells but is neither fastest nor universally lowest.

\begin{table}[H]
\centering
\small
\caption{Full direction ablation: 21 and 22 denote 20221121 and 20221122; each suffix is $n_s$, and cells give ten-seed mean $L_2$ error (\%)/time (s).}
\label{tab:ablation-full}
\setlength{\tabcolsep}{3.5pt}
\begin{tabular}{l c c c c c c c}
\toprule
Direction & 21/3 & 21/5 & 21/7 & 22/3 & 22/5 & 22/7 & Mean \\
\midrule
Spatial & \textbf{20.69/392} & \textbf{15.06/583} & \textbf{14.07/519} & \textbf{19.70/438} & 17.17/426 & 15.41/484 & \textbf{17.02/474} \\
Temporal & 21.48/380 & 16.21/329 & 14.75/328 & 21.66/330 & 17.65/331 & \textbf{15.25/338} & 17.83/339 \\
Space--time & 20.89/901 & 15.25/931 & 14.12/1174 & 19.81/907 & \textbf{17.09/1029} & 15.28/1043 & 17.07/998 \\
\bottomrule
\end{tabular}
\end{table}

\end{document}